\documentclass[10pt,twocolumn,letterpaper]{article}

\usepackage[pagenumbers]{cvpr} 

\usepackage[dvipsnames]{xcolor}

\usepackage{booktabs}
\usepackage{tabularx}
\usepackage{makecell}

\usepackage{xcolor}
\usepackage{colortbl}

\definecolor{orangeHighlight}{RGB}{255,160,0}

\definecolor{cvprblue}{rgb}{0.21,0.49,0.74}
\usepackage[pagebackref,breaklinks,colorlinks,citecolor=cvprblue]{hyperref}
\usepackage{booktabs, multirow, makecell, siunitx, xcolor, rotating}
\def\paperID{xxxx}
\def\confName{CVPR\xspace}
\def\confYear{2026\xspace}

\title{MVInverse: Feed-forward Multiview Inverse Rendering in Seconds}
\author{
{Xiangzuo Wu\textsuperscript{1} \quad
Chengwei Ren\textsuperscript{2} \quad
Jun Zhou\textsuperscript{1} \quad
Xiu Li\textsuperscript{1}\footnotemark[2] \quad
Yuan Liu\textsuperscript{2}\footnotemark[2]} \\
{\textsuperscript{1}Tsinghua University \quad
\textsuperscript{2}Hong Kong University of Science and Technology } \\
}

\newcommand{\best}{\cellcolor{orange!90}}
\newcommand{\sbest}{\cellcolor{orange!50}}
\newcommand{\tbest}{\cellcolor{orange!20}}

\newcommand{\tdown}{$\downarrow$}
\newcommand{\tup}{$\uparrow$}
\definecolor{citecolor}{HTML}{0071bc}
\definecolor{frontcolor}{HTML}{325ea5}
\definecolor{backcolor}{HTML}{a58b77}
\definecolor{sidecolor}{HTML}{10768c}
\definecolor{skincolor}{HTML}{dcb7b7}
\definecolor{darkred}{rgb}{0.6, 0.1, 0.05}
\definecolor{DeltaColor}{rgb}{0.039,0.73,0.71}
\definecolor{SigmaColor}{rgb}{0.98,0.45,0.0}
\definecolor{AlphaColor}{rgb}{0,0,0.8}
\definecolor{BetaColor}{rgb}{0.8,0,0.8}
\definecolor{GammaColor}{rgb}{0.514,0.34,0.224}
\definecolor{EpsilonColor}{rgb}{0.353,0.725,0.906}
\definecolor{PurpleColor}{HTML}{8B008B}
\definecolor{BadColor}{HTML}{C0392B}
\definecolor{OrangeColor}{rgb}{0.914,0.541,0.0.141}
\definecolor{GreenColor}{HTML}{00ab41}
\definecolor{RedColor}{rgb}{0.949,0.275, 0.224}
\definecolor{LightCyan}{rgb}{0.88,1,1}
\definecolor{Gray}{gray}{0.85}
\definecolor{LightGray}{gray}{0.70}

\definecolor{greenprior}{HTML}{34a853}
\definecolor{redprior}{HTML}{ea4335}
\definecolor{blueprior}{HTML}{4285f4}

\definecolor{bestcolor}{rgb}{1, 0.5, 0.25}
\definecolor{secondbestcolor}{rgb}{1, 0.8, 0.5}

\begin{document}


\twocolumn[{%
\renewcommand\twocolumn[1][]{#1}%
\maketitle
\centering
\includegraphics[width=\linewidth]{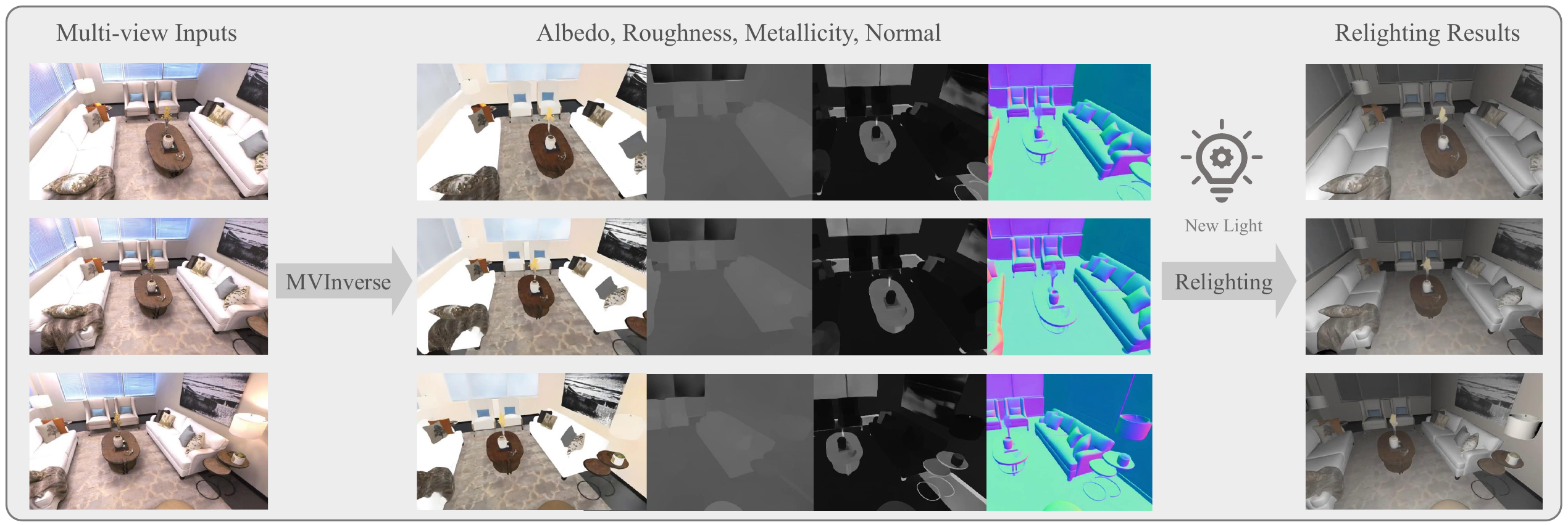}
\captionof{figure}{ 
We present \textbf{MVInverse}, a feed-forward multi-view inverse rendering framework that jointly recovers consistent geometry and material properties from input images or videos. With the recovered properties, realistic relighting under novel illumination can be achieved within seconds.
}
\label{fig:teaser}
}]

\renewcommand{\thefootnote}{\fnsymbol{footnote}}
\footnotetext[2]{Corresponding author}

\begin{abstract}
Multi-view inverse rendering aims to recover geometry, materials, and illumination consistently across multiple viewpoints. When applied to multi-view images, existing single-view approaches often ignore cross-view relationships, leading to inconsistent results. In contrast, multi-view optimization methods rely on slow differentiable rendering and per-scene refinement, making them computationally expensive and hard to scale. To address these limitations, we introduce a feed-forward multi-view inverse rendering framework that directly predicts spatially varying albedo, metallic, roughness, diffuse shading, and surface normals from sequences of RGB images. By alternating attention across views, our model captures both intra-view long-range lighting interactions and inter-view material consistency, enabling coherent scene-level reasoning within a single forward pass. Due to the scarcity of real-world training data, models trained on existing synthetic datasets often struggle to generalize to real-world scenes. To overcome this limitation, we propose a consistency-based finetuning strategy that leverages unlabeled real-world videos to enhance both multi-view coherence and robustness under in-the-wild conditions. Extensive experiments on benchmark datasets demonstrate that our method achieves state-of-the-art performance in terms of multi-view consistency, material and normal estimation quality, and generalization to real-world imagery.   Code, models and more results are available
at \url{https://maddog241.github.io/mvinverse-page/}
\end{abstract}    
\section{Introduction}
\label{sec:intro}

\begin{figure}[t!]
\centering
\includegraphics[width=0.95\linewidth]{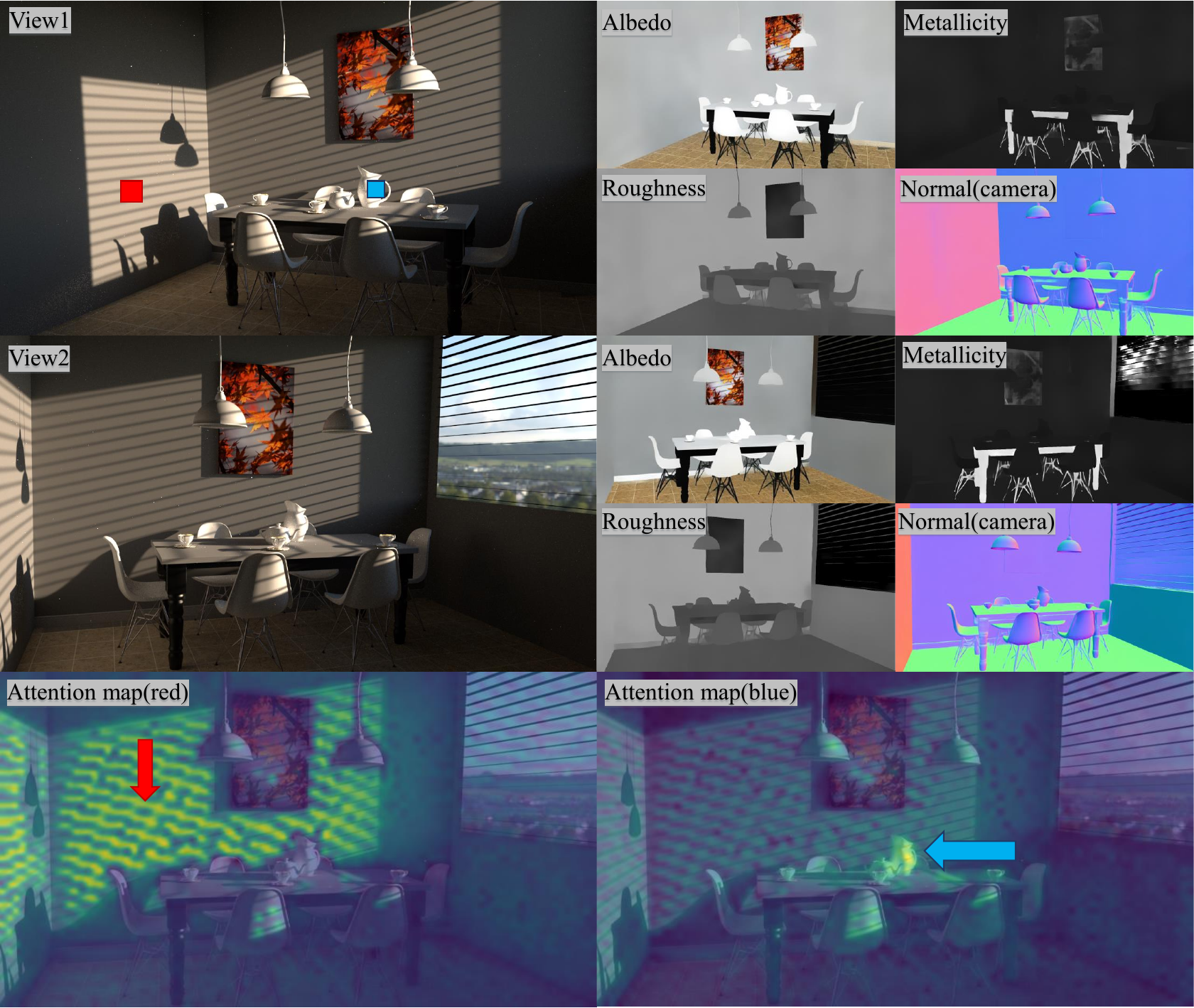}
\caption{\textbf{Alternating Attention}. 
Given the red and blue query patches in the first view, we visualize the corresponding attention heatmaps in the second view (bottom row) to illustrate the effectiveness of our model design. 
\textcolor{red}{Red:} the model captures long-range lighting interactions across spatially distant regions. 
\textcolor{blue}{Blue:} the model maintains cross-view consistency by correctly associating corresponding surface regions under viewpoint changes.}
\label{fig:motivation}
\vspace{-4mm}
\end{figure}

Inverse rendering, a fundamental task in computer vision and graphics, aims to recover intrinsic scene properties—including albedo, roughness, and metallicity—from visual observations. This capability is pivotal across numerous practical domains: in film and animation production, it enables efficient digital asset creation by reconstructing physically accurate materials from real-world captures~\cite{xiong2025texgaussian, xiang2025structured, zhao2025hunyuan3d}; in augmented reality (AR) and virtual reality (VR), it facilitates seamless integration of virtual objects with real environments by matching their material properties to the scene~\cite{ren2025mv, careaga2025physically, peng2025interactive}; and in robotics, it enhances scene understanding for tasks like object manipulation by inferring physical characteristics of surfaces~\cite{zhai2024physical, cai2024gic, xu2025gaussianproperty}. Beyond these applications, inverse rendering serves as a cornerstone for advancing physically based rendering (PBR) pipelines, bridging the gap between image-based perception and 3D scene reconstruction.

Despite its significance, existing inverse rendering methods face critical limitations that hinder their practical deployment. First, the majority of approaches rely on iterative optimization frameworks~\cite{azinovic2019inverse, nimier2021material, wu2023factorized, yu2023milo, srinivasan2021nerv, boss2021nerd, knodt2021neural, zhang2021nerfactor, gao2024relightable, jiang2024gaussianshader, liang2024gs}, which often yield high-fidelity results but suffer from prohibitive computational costs, requiring minutes to hours of processing per scene, making them infeasible for real-time or large-scale applications. Second, while some single-view intrinsic decomposition methods~\cite{li2018learning, li2018materials, li2020inverse, li2022physically, zhu2022learning,careagaColorful} achieve fast inference by leveraging feed-forward networks, their extension to multi-view settings is plagued by multi-view inconsistency: material predictions for the same 3D region across different viewpoints often diverge, leading to unrealistic artifacts in reconstructed scenes. Third, recent diffusion-based renderers have demonstrated promising results for image and video inverse rendering~\cite{zeng2024rgb, kocsis2024intrinsic, liang2025diffusion}, but their inherent generative nature and iterative sampling process render them computationally expensive, inheriting the speed limitations of optimization-based methods. Thus, developing an inverse rendering solution that balances speed and accuracy remains an open and pressing challenge.

Our work is motivated by the remarkable recent progress in feed-forward 3D reconstruction, where vision transformer (ViT)-based models have revolutionized the field by replacing lengthy optimization pipelines with direct, end-to-end prediction~\cite{wang2025pi, wang2025vggt, keetha2025mapanything, liu2025worldmirror}. Multi-view inverse rendering shares a key similarity with these reconstruction methods: both traditionally depend on repeated rendering iterations to solve for intrinsic properties (materials for inverse rendering, geometry for reconstruction). This parallel insight suggests that integrating the feed-forward paradigm into inverse rendering is a natural and unexplored direction. By leveraging the efficiency of transformer-based architectures to directly map multi-view inputs to material properties, we can potentially overcome the speed bottlenecks of optimization and diffusion-based methods while addressing the consistency issues of monocular approaches.

Building on this motivation, we propose \textbf{MVInverse}, a novel feed-forward model that predicts albedo, roughness, and metallicity along with camera-space normals and diffuse shading maps for all input views in just a few seconds. Inspired by recent state-of-the-art feed-forward 3D reconstruction architectures~\cite{wang2025vggt, wang2025pi}, we design a transformer backbone with alternating global and frame-wise attention: the global attention explicitly associates corresponding 3D regions across different viewpoints, while the frame-wise attention captures intra-frame interactions in individual images. This alternating-attention design extracts efficient, viewpoint-aligned features that serve as a robust foundation for subsequent material and normal prediction. As illustrated in Figure~\ref{fig:motivation}, attention visualizations reveal that our model effectively captures both long-range lighting dependencies (red patch) and inter-frame consistency (blue patch), highlighting the dual capability of the alternating attention mechanism. 

A further challenge in inverse rendering is the scarcity of high-quality real-world training data. Most existing methods are trained on synthetic datasets~\cite{roberts2021hypersim, zhu2022learning, wang2022high, zheng2020structured3d, collins2022abo, li2023matrixcity, li2018cgintrinsics}, leading to poor generalization to real-world scenarios, particularly the emergence of flickering artifacts, where material properties of the same 3D region fluctuate inconsistently across views or frames. To mitigate this, we adopt a two-stage training strategy: we first pre-train our model on large-scale synthetic datasets to learn general material priors, then introduce a self-supervised consistency constraint during fine-tuning on real-world data. This constraint enforces that material predictions for the same 3D point, as observed from different viewpoints, remain consistent, significantly improving the model’s generalization to real environments and eliminating flickering artifacts.

We validate our approach across common benchmark datasets, demonstrating strong single-view inverse rendering quality, multi-view material consistency, accurate normal prediction, and stable temporal performance on videos. Moreover, by combining our method with existing feed-forward 3D reconstruction pipelines, we can obtain textured point clouds that enable various downstream applications-such as video relighting and view-consistent material editing-in only a few seconds. This efficient end-to-end pipeline greatly lowers the barrier for practical deployment, paving the way for highly scalable, data-driven material understanding. 
\section{Related Work}

\paragraph{Single-view Inverse Rendering}

Inverse rendering aims to decompose a scene into geometry, material properties and illumination~\cite{barrow1978recovering}. This is an inherently ill-posed task as multiple combinations of these properties can produce the same image. Early approaches relied on handcrafted priors and physical heuristics to constrain the problem ~\cite{land1971lightness, barrow1978recovering, finlayson2004intrinsic, shen2011intrinsic, shen2008intrinsic, grosse2009ground, bell2014intrinsic, barron2014shape,  zhang2021unsupervised, zhao2012closed}, which were often limited by the applicability of their underlying assumptions to diverse real-world scenes.
The advent of deep learning has shifted the paradigm towards data-driven techniques. Modern methods often frame inverse rendering as an image-to-image translation problem, training dense prediction neural networks on large-scale synthetic datasets with ground truth decompositions~\cite{shi2017learning, li2018learning, li2018materials, ma2018single, sengupta2019neural, yu2019inverserendernet, sang2020single, li2020inverse, wang2021learning, zhu2022irisformer, li2022physically, zhu2022learning, wimbauer2022rendering, careagaIntrinsic, careagaColorful, zhang2024learning, zhan2022gmlight, philip2021free}. These learning-based approaches can jointly estimate spatially-varying bidirectional reflectance distribution functions (SVBRDFs), complex lighting, and surface normals. 
More recently, generative models, particularly diffusion models, have been employed to address the problem in a formulation of conditional image generation~\cite{du2023generative, liang2024photorealistic, phongthawee2024diffusionlight, liang2025luxdit, zeng2024rgb, kocsis2024intrinsic, liang2025diffusion, djeghim2025sail}. These models leverage powerful priors learned from large-scale image distributions to synthesize physically plausible decompositions. However, a fundamental limitation of all single-view methods is their inability to enforce geometric and material consistency across different viewpoints of the same scene. In contrast, our proposed method leverages multiple views simultaneously to resolve this ambiguity and ensure a coherent scene decomposition.

\paragraph{Multi-view Inverse Rendering}

Multi-view inverse rendering leverages information from multiple observations of a scene to enable more accurate recovery of geometry, materials, and illumination. Early multi-view methods often relied on explicit geometry and heavy optimization to jointly refine these factors~\cite{azinovic2019inverse, nimier2021material, wu2023factorized, yu2023milo, Mitsuba3}, achieving physically accurate results through iterative optimization.
With the rise of neural scene representations, NeRF~\cite{mildenhall2021nerf} and NeuS~\cite{wang2021neus}-based approaches~\cite{srinivasan2021nerv, boss2021nerd, knodt2021neural, zhang2021nerfactor, boss2021neural, yang2022ps, yao2022neilf, chen2022tracing, boss2022samurai, attal2024flash, zhang2022iron, liu2023nero, yang2024robir, wang2024inverse, cai2024pbir} have become a dominant paradigm. These methods model radiance and reflectance within implicit neural fields and typically optimize network parameters per scene to disentangle view-dependent appearance and incident illumination.
More recently, 3D Gaussian Splatting~\cite{kerbl20233d} has been adapted for inverse rendering~\cite{gao2024relightable, jiang2024gaussianshader, liang2024gs, shi2025gir, wu2024deferredgs, chung2024differentiable, zhu2024gs, guo2024prtgs, ye2024progressive, dihlmann2024subsurface, bi2024gs3, chen2024gi, lai2025glossygs, ye2025geosplatting, gu2025irgs}, offering real-time differentiable rendering and efficient geometry representation with some feed-forward 3D geometry reconstruction~\cite{wang2025vggt,wang2025pi,lu2025align3r, gao2025more, fang2025dens3r}.
In addition, several hybrid methods~\cite{lin2025iris, litman2025materialfusion, chen2024intrinsicanything, lyu2023diffusion} adopt an estimation-then-optimization strategy, where learning-based predictions are refined through per-scene optimization. Despite recent progress, most existing approaches still rely on scene-specific optimization or fine-tuning to achieve material and lighting consistency. Although learning-based systems such as MAIR and MAIR++~\cite{choi2023mair, choi2025mair++} consider multi-view settings without optimization, their applicability remains limited to relatively constrained viewpoints.
In contrast, our method performs multi-view inverse rendering in a single feed-forward pass, producing coherent scene decomposition even under large camera motions.
\section{Method}

\begin{figure*}[t!]
\centering
\includegraphics[width=0.95\linewidth]{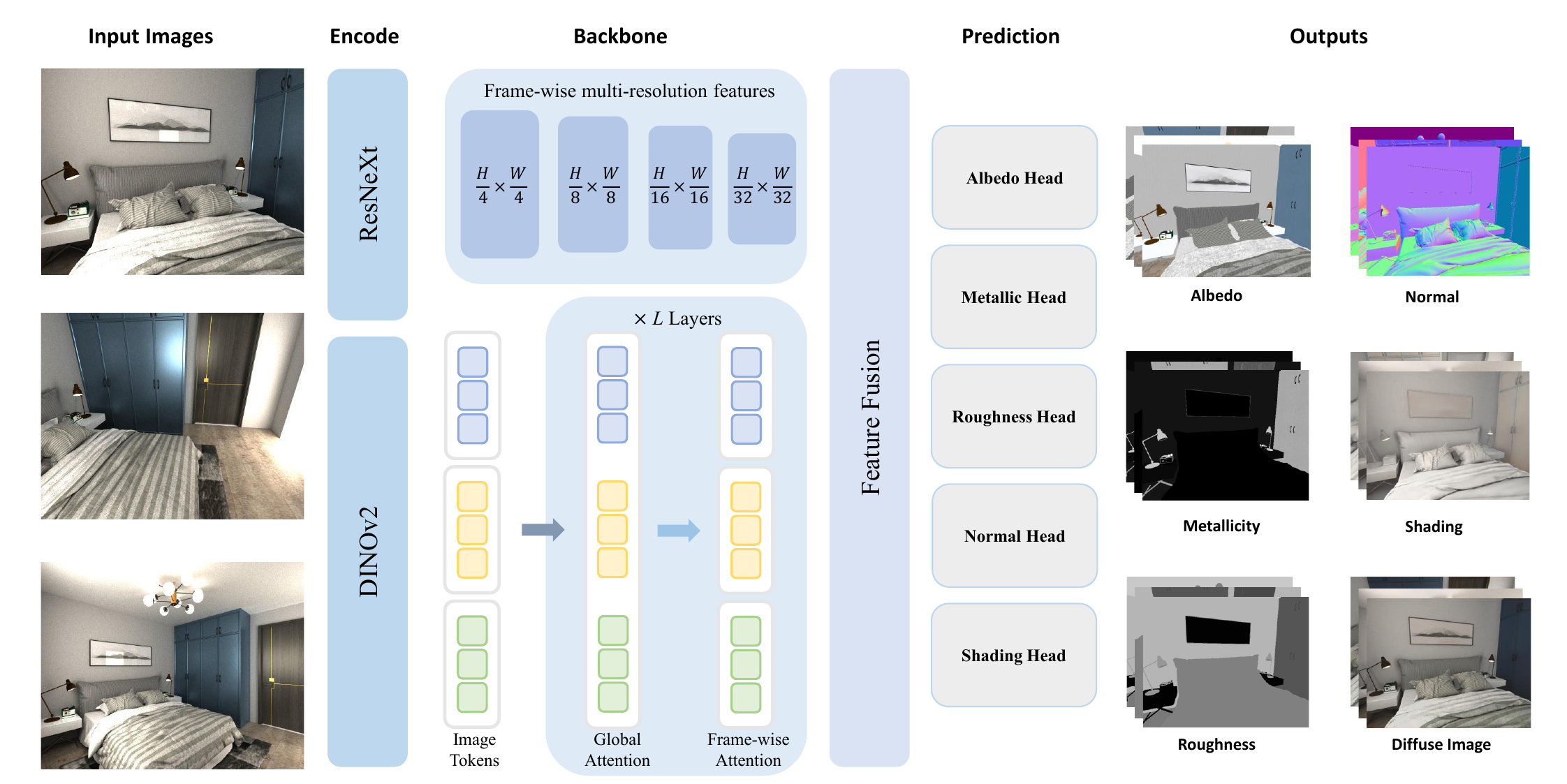}
\caption{\textbf{Framework of MVInverse.}  
Given an input image sequence, our framework first encodes each frame into patch tokens via DINOv2~\cite{oquab2023dinov2}, where alternating global–frame attention enables cross-view feature aggregation.
Meanwhile, a frame-wise ResNeXt~\cite{xie2017aggregated} encoder provides multi-resolution convolutional features to preserve fine spatial details.
The two feature streams are fused in a DPT-style prediction head to produce pixel-aligned intrinsic maps, including albedo, metallic, roughness, normal, and shading.
The diffuse image is obtained as the product of albedo and diffuse shading.
}
\label{fig:method}
\end{figure*}

In this section, we introduce MVInverse, a feed-forward inverse rendering framwork which directly predicts multiple intrinsic properties given a sequence of images. We firstly introduce the problem definition and our notations in Sec~\ref{sec:definition}, followed by our model architecture and design choices in Sec~\ref{sec:network}. In Sec~\ref{sec:consistency}, we introduce a consistency finetuning scheme, improving the model's generalization capability in real-world scenes.

\subsection{Problem Definition and Notation}
\label{sec:definition}

Given a sequence of $N$ images $S = (I_{1}, ..., I_{N}), I_i \in \mathbb{R}^{H \times W \times 3}$, the network $f$ aims to map the sequence to a corresponding set of intrinsic images: 

\begin{equation}
    f((I_i)_{i=1}^{N}) = (A_i, M_i, R_i, N_i, D_i)_{i=1}^N,
\end{equation}
where $A_i \in [0, 1]^{H \times W \times 3}$ denotes the albedo map, representing the surface's base reflectance color. 
$M_i \in [0,1]^{H \times W}$ is the metallic map, and 
$R_i \in [0,1]^{H \times W}$ represents the roughness map. 
$N_i \in [-1,1]^{H \times W \times 3}$ is the surface normal map, defined in the camera coordinate system.
$D_i \in [0,1]^{H \times W \times 3}$ denotes the diffuse shading map, capturing the total incoming radiance over the upper hemisphere at each surface point, which contributes to the diffuse shading.

\subsection{MVInverse}
\label{sec:network}

\paragraph{Feature Backbone}
Following recent advances in feed-forward 3D reconstruction~\cite{wang2025vggt, wang2025pi}, we adopt a similar architecture tailored for the multi-view inverse rendering setting, as illustrated in Figure~\ref{fig:method}. 

We use DINOv2~\cite{oquab2023dinov2} as the visual feature encoder due to its strong generalization capability and ability to extract semantically meaningful high-level representations. Each input image $I_i$ is converted into a set of visual tokens: $X_i = \mathrm{DINOv2}(I_i), \quad X_i \in \mathbb{R}^{T \times C}$, where $T$ denotes the number of tokens and $C$ is the feature dimensionality.

To jointly exploit intra-view appearance cues and cross-view intrinsic consistency, we adopt a permutation-equivariant alternating-attention transformer backbone from Pi3~\cite{wang2025pi}. The backbone alternates between two complementary attention operations:

\begin{itemize}
    \item \textbf{Frame-wise self-attention}, which performs attention \emph{within a single image} and refines token representations by capturing long-range spatial dependencies and local semantic structures.
    \item \textbf{Global self-attention}, which performs attention across all input images and allows tokens from different views to mutually reference and reinforce each other. This mechanism implicitly associates tokens that observe the same underlying 3D surface region, without requiring camera pose supervision.
\end{itemize}

Through progressive alternation between these two attention operations, the network injects rich intra-view contextual reasoning and globally consistent cross-view information into each token. This yields feature representations that are structurally coherent and view-consistent, forming a reliable basis for multi-view inverse rendering.

\paragraph{Intrinsic Heads}
\label{sec:intrinsic_head}

Given the rich multi-view latent representation, MVInverse predicts intrinsic properties for every input image through five dedicated prediction heads:
\[
(A_i, M_i, R_i, N_i, D_i) = \mathrm{Head}(X_i).
\]
Each head is implemented as a DPT-style dense prediction module~\cite{ranftl2021vision}, producing pixel-aligned predictions at the input resolution.

However, unlike depth and semantic segmentation tasks, where DPT is commonly used, inverse rendering requires much stronger spatial detail preservation. Empirically, we find that using the DPT head alone often leads to overly smoothed and blurry material maps, particularly in albedo, where high-frequency texture patterns must be recovered accurately.
To overcome this limitation, we introduce an \textit{auxiliary multi-resolution convolutional encoder} that extracts local high-frequency structures from the input image and injects them into the prediction head via feature fusion. The extracted fine-grained details are skip-connected to the prediction heads to preserve texture sharpness while avoiding unnecessary blurring. As shown in Figure~\ref{fig:ablation_res}, this DPT-hybrid-like~\cite{ranftl2021vision} architecture substantially improves spatial fidelity and restores fine-scale reflectance variations.

Each intrinsic property is supervised with a tailored loss. 
For albedo, metallic, roughness, and diffuse shading, we follow~\cite{li2018cgintrinsics, li2018megadepth, careagaColorful}
to use mean-squared error (MSE) loss and multi-scale gradient (MSG) loss:
\begin{equation}
    \mathcal{L}_{mse}(P) = \frac{1}{N} \sum_{i=1}^{N} (P_i - P_i^{*})^{2},
\end{equation}
\begin{equation}
    \mathcal{L}_{msg}(P) = \frac{1}{NM} \sum_{i=1}^{N} \sum_{l=1}^{M}
        \left( \nabla P_{i,l} - \nabla P^{*}_{i,l} \right)^{2},
\end{equation}

where $P \in \{A, M, R, D\}$ represents the predicted intrinsic map (albedo, metallic, roughness, or diffuse shading), $P^{*}$ is the corresponding ground-truth supervision, 
and $\nabla P_{i,l}$ denotes the spatial gradient of $P$ at scale $l$. 
Note that a scale-invariant MSE loss is applied for albedo prediction, with details provided in the supplementary materials.

Surface normals are trained using a cosine similarity loss, enforcing correct orientation rather than raw vector magnitude: 

\begin{equation}
    \mathcal{L}_{normal}(N_i) = 1 - \langle \hat{N}_i, N_i \rangle.
\end{equation}

\paragraph{Training}
Our model is trained in an end-to-end manner by optimizing a composite loss function that simultaneously constrains multiple material properties, including albedo, metallicity, roughness, normals, and shading. Each individual loss term is weighted in the overall objective, enabling the model to learn these physical attributes jointly.

\begin{align}
\mathcal{L} &= \lambda_{\text{albedo}} \, \mathcal{L}_{\text{albedo}}
             + \lambda_{\text{metallic}} \, \mathcal{L}_{\text{metallic}}
             + \lambda_{\text{roughness}} \, \mathcal{L}_{\text{roughness}} \nonumber \\
            &\quad + \lambda_{\text{normal}} \, \mathcal{L}_{\text{normal}}
            + \lambda_{\text{shading}} \, \mathcal{L}_{\text{shading}} 
\end{align}
where $\lambda_{\text{*}}$ denotes the weighting factor for each component, and $\mathcal{L}_{\text{*}}$ corresponds to the loss of the respective material property. In practice, we set all weighting factors $\lambda_{\text{*}}=1$. 

Our model is trained on a diverse collection of datasets to ensure robustness across various environments. Specifically, we use Hypersim~\cite{roberts2021hypersim}, Interiorverse~\cite{zhu2022learning}, PRID~\cite{wang2022high}, Structured3D~\cite{zheng2020structured3d}, Amazon-Berkeley Objects~\cite{collins2022abo}, MatrixCity~\cite{li2023matrixcity}, and CGIntrinsics~\cite{li2018cgintrinsics} as the primary training sources.
Since most of these datasets contain synthetic indoor or object-centric scenes, limiting generalization to outdoor scenes, we additionally use DiffusionRenderer~\cite{liang2025diffusion} to generate pseudo albedo labels for 1,118 real-world videos from the Sekai~\cite{li2025sekai}-Drone dataset. Additional training details and dataset specifications can be found in the supplementary material. 

\subsection{Consistency Finetuning}
\label{sec:consistency}

After pretraining, the network already produces view-consistent material predictions in synthetic scenes. However, when applied to real-world videos, we observe noticeable temporal flickering on the same 3D regions, resulting in poor visual stability. We resort the issue to the lack of ground-truth supervision in real-world scenes.

\begin{figure}
\centering
\includegraphics[width=\linewidth]{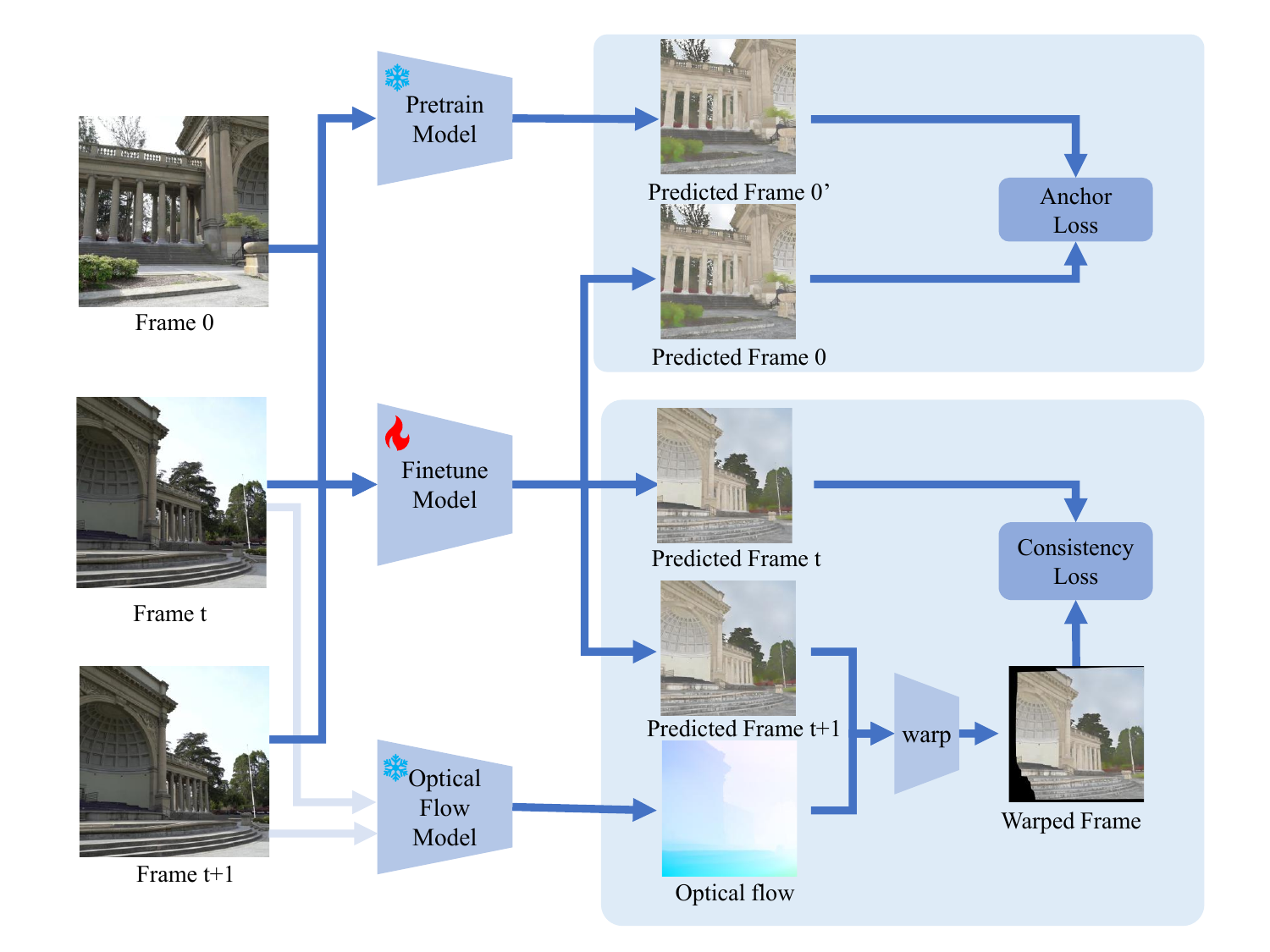}
\caption{\textbf{Consistency finetuning on real videos.} 
Frames are fed into the finetuned model to produce predictions. Optical flow warps the prediction of frame $t+1$ to frame $t$, and a consistency loss enforces temporal stability. An anchor loss on frame $0$, using a pretrained model's prediction as reference, prevents solution collapse.
}
\label{fig:consistency}
\end{figure}

To address this, we adopt a two-stage training scheme. After large-scale pretraining on synthetic data, we perform self-supervised finetuning on real-world videos to enhance temporal consistency, as illustrated in Figure~\ref{fig:consistency}. Given a sequence of input frames ($I_0, I_t, I_{t+1}$), the finetuned model generates material predictions ($\hat{M}_0, \hat{M}_t, \hat{M}_{t+1}$). To enforce temporal coherence, we estimate dense optical flow $F_{t+1 \rightarrow t}$ between frames $I_{t+1}$ and $I_t$ using an off-the-shelf optical flow network~\cite{teed2020raft}. The predicted material map $\hat{M}_{t+1}$ is warped according to $F_{t+1 \rightarrow t}$ to obtain $\hat{M}_{t+1 \rightarrow t}^{warp}$. A consistency loss is computed between $\hat{M}_{t+1 \rightarrow t}^{warp}$ and $\hat{M}_t$, encouraging adjacent predictions to remain consistent under motion.

However, applying only consistency supervision may lead to trivial or collapsed solutions where the model learns to generate temporally smooth but semantically meaningless outputs. To prevent this, we introduce an anchor loss at frame 0. A pretrained model produces a reference material map $\hat{M}_0^{pret}$, and the finetuned model is required to match this prediction at the same frame. This anchor loss preserves the pretrained model's predictive ability while still allowing finetuning to improve temporal stability.
\begin{equation}
\mathcal{L}_{\text{finetune}} 
= \lambda_{\text{anchor}}\,\|\hat{M}_0 - \hat{M}_0^{\text{pred}}\|_2^2 
+ \|\hat{M}_t - \hat{M}_{t+1 \rightarrow t}^{\text{warp}}\|_2^2.
\end{equation}
Here, we set $\lambda_{\text{anchor}}=0.1$ to balance the contributions of the anchor and consistency losses.
Together, these losses guide the model to learn video-stable material predictions without sacrificing the fidelity obtained from pretraining.

\begin{table}[tb]
\footnotesize
\centering
\caption{\textbf{Quantitative evaluation of single-view inverse rendering on \textbf{Interiorverse~\cite{zhu2022learning}}.} 
* We do inference in the image mode. 
Best / 2nd / 3rd per column are highlighted.}
\label{tab:singleview_material}
\resizebox{\columnwidth}{!}{
\begin{tabular}{l|ccc|c|c}
\toprule
 & \multicolumn{3}{c|}{\textbf{Albedo}} & \textbf{Metallic} & \textbf{Roughness} \\
\cmidrule(lr){2-4} \cmidrule(lr){5-5} \cmidrule(lr){6-6}
 & PSNR $\uparrow$ & SSIM $\uparrow$ & LPIPS $\downarrow$
 & RMSE $\downarrow$ & RMSE $\downarrow$ \\
\midrule
IIW~\cite{bell2014intrinsic} & 9.7 & 0.62 & 0.47 & -- & -- \\
RGB$\leftrightarrow$X~\cite{zeng2024rgb} & 16.4 & 0.78 & \tbest{0.19} & 0.44 & 0.38 \\
ColorfulDiffuse~\cite{careagaColorful} & 15.2  & 0.76  & 0.35  & -- & -- \\
IntrinsicImageDiffusion~\cite{kocsis2024intrinsic} & \tbest{17.4} & \tbest{0.80} & \sbest{0.22} & \sbest{0.21} & \sbest{0.26} \\
DiffusionRenderer~\cite{liang2025diffusion}* & \sbest{21.9} & \sbest{0.87} & \sbest{0.17} & \tbest{0.28} & \tbest{0.35} \\
Ours & \best{23.0} & \best{0.92} & \best{0.09} & \best{0.14} & \best{0.17} \\
\bottomrule
\end{tabular}
}
\vspace{-4mm}
\end{table}

\section{Experiments}
We conduct extensive experiments to evaluate the proposed feed-forward multi-view inverse rendering framework. Our evaluation focuses on three core aspects: (1) single-view material estimation quality, (2) multi-view material consistency, and (3) normal estimation quality. Through both quantitative and qualitative comparisons, we demonstrate that our method achieves efficient and consistent scene-level inverse rendering without per-scene optimization.

\subsection{Single-view Material Estimation}

\begin{figure*}[t!]
\centering
\includegraphics[width=\linewidth]{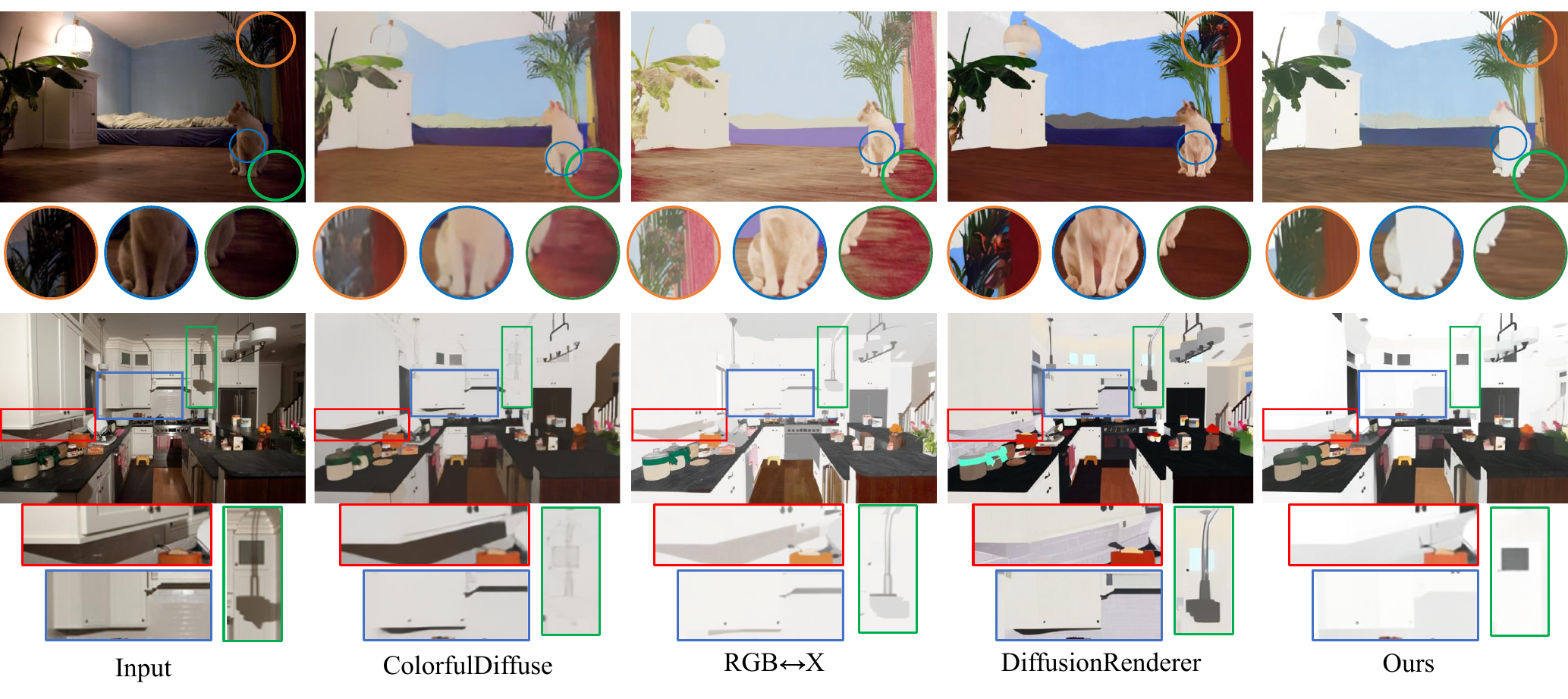}
\vspace{-7mm}
\caption{\textbf{Single-view Albedo Comparison.} In the top row, \textcolor{green}{green} circles highlight regions where lighting is not removed; \textcolor{blue}{blue} circles highlights regions where shadow still exists;
\textcolor{orange}{orange} circles demonstrate an unnatural red patch inside the leaves. In the bottom row, \textcolor{red}{red} and \textcolor{blue}{blue} regions indicate areas where small shadows are not removed; \textcolor{green}{green} regions highlight shadows cast by the pendant light on the cabinet that are not recognized by other models. In contrast, our model handles these cases well.
}
\label{fig:singleview_qualitative}
\end{figure*}
We first evaluate single-view material estimation on the test set of Interiorverse~\cite{zhu2022learning} with 2672 image-material pairs. 
We evaluate albedo predictions using PSNR, SSIM, and LPIPS, and roughness/metallic using RMSE. For comparison, we include state-of-the-art intrinsic decomposition and inverse rendering methods: DiffusionRenderer~\cite{liang2025diffusion}, ColorfulDiffuse~\cite{careagaColorful}, RGB$\leftrightarrow$X~\cite{zeng2024rgb}, IntrinsicImageDiffusion~\cite{kocsis2024intrinsic}, and earlier approaches~\cite{bell2014intrinsic}. Quantitative results in Table~\ref{tab:singleview_material} show that our method outperforms all baselines across all evaluated metrics. This demonstrates that our approach not only achieves higher accuracy in albedo and material property estimation but also produces more perceptually faithful and visually consistent results

We further provide qualitative comparisons for zero-shot real-world single-view albedo estimation in Intrinsic Images in the Wild (IIW)~\cite{bell2014intrinsic} dataset, as shown in Figure~\ref{fig:singleview_qualitative}. In the top row, regions highlighted in green indicate areas where RGB$\leftrightarrow$X~\cite{zeng2024rgb} and ColorfulDiffuse~\cite{careagaColorful} fail to correctly remove lighting effects, while blue regions show that these methods do not remove the cat’s shadow. Additionally, orange regions illustrate that DiffusionRenderer~\cite{liang2025diffusion} generates unnatural color artifacts, such as a red patch in the leaves, and exhibits suboptimal overall color tone.
In the bottom row, red and blue regions correspond to small shadows under the cabinet that are not removed by existing methods. Green regions highlight the shadows of pendant light on the cabinet that other models fail to recognize. In contrast, our method effectively eliminates these shadows, demonstrating superior capability in both shadow removal and albedo recovery under challenging lighting conditions.

\subsection{Multi-view Material Consistency}

\begin{figure*}[t!]
\centering
\includegraphics[width=\linewidth]{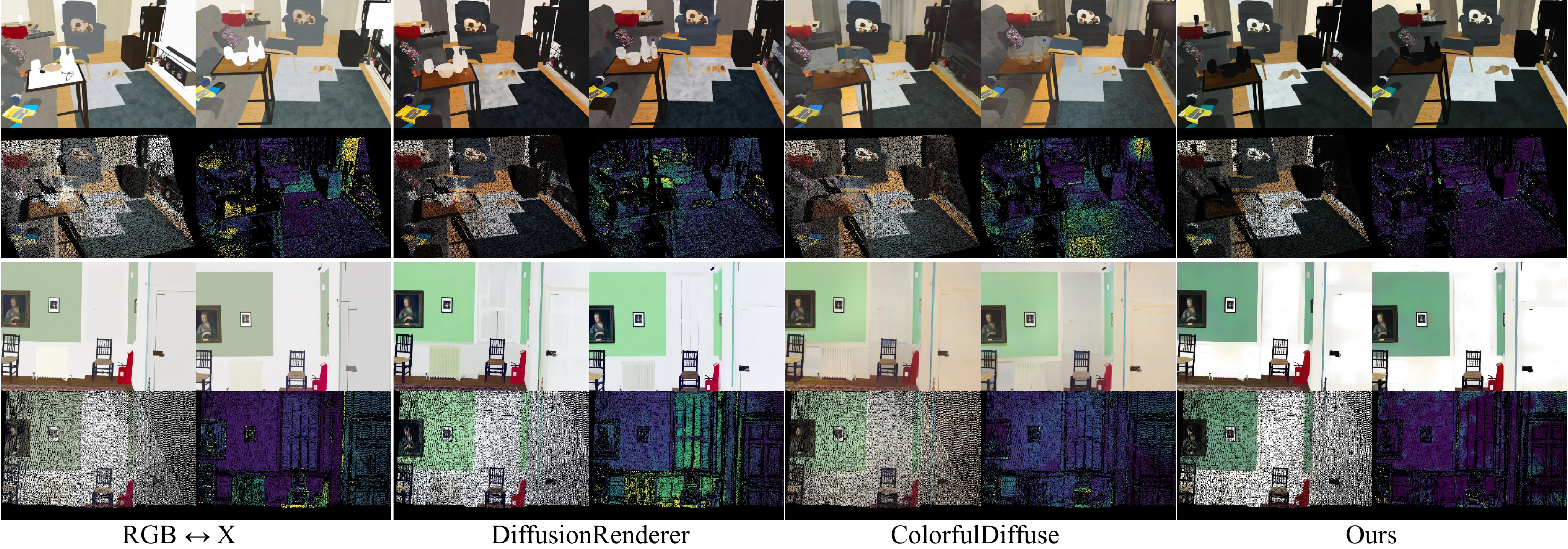}
\vspace{-6mm}
\caption{\textbf{Multi-view Albedo Comparison}
We compare our method with existing approaches under multi-view settings. 
For each scene, the top-left and top-right subfigures show the albedo predictions from \textit{view 1} and \textit{view 2}, respectively. 
The bottom-left subfigure shows the result of warping the prediction from \textit{view 2} to \textit{view 1}, and the bottom-right subfigure visualizes the per-pixel difference between the two views. 
Our method demonstrates higher cross-view consistency and fewer artifacts across viewpoints.
}
\label{fig:multiview_qualitative}
\vspace{-4mm}
\end{figure*}
\begin{table}[tb]
\footnotesize
\centering
\caption{RMSE for \textbf{Albedo}, \textbf{Metallic}, and \textbf{Roughness}. Results are evaluated on multi-view images in Hypersim~\cite{roberts2021hypersim}.}
\label{tab:multiview_material}
\begin{tabular}{lccc}
\toprule
\multirow{2}{*}{Method} & \multicolumn{3}{c}{RMSE\tdown} \\
\cmidrule(lr){2-4}
 & Albedo & Metallic & Roughness \\
\midrule
RGB$\leftrightarrow$X~\cite{zeng2024rgb} & 0.1007 & 0.2941 & 0.1065 \\
ColorfulDiffuse~\cite{careagaColorful} & 0.1001 & - & - \\
IntrinsicImageDiffusion~\cite{kocsis2024intrinsic} & \sbest{0.0723} & \sbest{0.0720} & \sbest{0.0630} \\
DiffusionRenderer~\cite{liang2025diffusion} & \tbest{0.0826} & \tbest{0.0731} & \tbest{0.0995} \\
Ours & \best{0.0494} & \best{0.0587} & \best{0.0235} \\
\bottomrule
\end{tabular}
\vspace{-4mm}
\end{table}

In this section, we compare our model’s multi-view consistency with existing methods. Since there is no established protocol for evaluating multi-view material consistency, we design our own evaluation scheme. We select the last five scenes from the Hypersim~\cite{roberts2021hypersim} test split. Note thta none of the compared methods have seen these five test scenes during training. For each scene, we perform inference on the first 10 views in all camera trajectories. Using known depth maps and camera poses, predicted material maps (albedo, roughness, metallic) from one view are back-projected into 3D world coordinates and reprojected onto other views. We then compute RMSE between reprojected and directly predicted maps over overlapping regions to quantify cross-view consistency. Results in Table~\ref{tab:multiview_material} show that our method achieves the lowest RMSE across all material channels (albedo, roughness, and metallic), indicating superior cross-view consistency compared to existing approaches, which is an essential property for material acquisition in downstream applications.

We additionally provide qualitative comparisons on multi-view albedo consistency, as shown in Figure~\ref{fig:multiview_qualitative}. We select two views from the \textit{Room} scene in the Mip-NeRF 360~\cite{barron2022mipnerf360} dataset and \textit{Drjohnson} scene in Deep Blending~\cite{DeepBlending2018} dataset and predict their albedo using each method. The second view is warped to the first view using camera poses and depth estimated by Pi3~\cite{wang2025pi}. We then compute and visualize an error map between the predicted albedo of the first view and the warped albedo from the second view, where brighter pixels indicate larger inconsistencies between views.

As shown in Figure~\ref{fig:multiview_qualitative}, our model produces the least amount of bright pixels in the error map, indicating significantly lower cross-view discrepancies. This demonstrates that our method maintains strong multi-view albedo consistency, while other approaches exhibit noticeable view-dependent variations. 

\subsection{Normal Estimation}

\begin{table*}[thbp]
\centering
\caption{\textbf{Quantitative evaluation on zero-shot surface normal estimation.} Comparison across five benchmarks. Best / 2nd / 3rd per column are highlighted from deep to light orange.}
\label{tab:normal_benchmark}
\resizebox{\textwidth}{!}{
\begin{tabular}{l|ccc|ccc|ccc|ccc|ccc}
\toprule
& \multicolumn{3}{c|}{NYUv2~\cite{silberman2012indoor}} 
& \multicolumn{3}{c|}{ScanNet~\cite{dai2017scannet}} 
& \multicolumn{3}{c|}{iBims-1~\cite{koch2018evaluation}} 
& \multicolumn{3}{c|}{Sintel~\cite{butler2012naturalistic}} 
& \multicolumn{3}{c}{OASIS~\cite{chen2020oasis}} \\
Method & mean \tdown & 11.25$^{\circ}$ \tup & 30$^{\circ}$ \tup
       & mean \tdown & 11.25$^{\circ}$ \tup & 30$^{\circ}$ \tup
       & mean \tdown & 11.25$^{\circ}$ \tup & 30$^{\circ}$ \tup
       & mean \tdown & 11.25$^{\circ}$ \tup & 30$^{\circ}$ \tup
       & mean \tdown & 11.25$^{\circ}$ \tup & 30$^{\circ}$ \tup \\
\midrule
Omnidata V2~\cite{kar20223d}      & 17.2 & 55.5 & 83.0  & 16.2 & 60.2 & 84.7  & 18.2 & 63.9 & 81.1  & 40.5 & 14.7 & 43.5  & 24.2 & 27.7 & \tbest{74.2} \\
DSINE~\cite{bae2024rethinking}          & \tbest{16.4} & 59.6 & \tbest{83.5}  & 16.2 & 61.0 & 84.4  & \tbest{17.1} & \sbest{67.4} & 82.3  & 34.9 & 21.5 & 52.7  & 24.4 & 28.8 & 72.0 \\
GeoWizard~\cite{fu2024geowizard}         & 18.9 & 50.7 & 81.5  & 17.4 & 53.8 & 83.5  & 19.3 & 63.0 & 80.3  & 40.3 & 12.3 & 43.5  & 25.2 & 23.4 & 68.1 \\
StableNormal~\cite{ye2024stablenormal}   & 18.6 & 53.5 & 81.7  & 17.1 & 57.4 & 84.1  & 18.2 & 65.0 & 82.4  & 36.7 & 14.1 & 50.7  & 26.5 & 23.5 & 68.7 \\
Diffusion-E2E-FT~\cite{garcia2025fine}  & 16.5 & \best{60.4} & 83.1  
                                         & \sbest{14.7} & \sbest{66.1} & \tbest{85.1} 
                                         & \sbest{16.1} & \best{69.7} & \sbest{83.9}  
                                         & \tbest{33.5} & \tbest{22.3} & \tbest{53.5}  
                                         & \tbest{23.2} & \tbest{29.4} & \sbest{74.5} \\
Lotus-D~\cite{he2024lotus}          & \sbest{16.2} & \tbest{59.8} & \sbest{83.9}  
                                     & \tbest{14.7} & \tbest{64.0} & \sbest{86.1}  
                                     & \tbest{17.1} & 66.4 & \tbest{83.0}  
                                     & \sbest{32.3} & \sbest{22.4} & \sbest{57.0}  
                                     & \best{22.3} & \best{31.8} & \best{76.1} \\
\textbf{Ours}              & \best{16.1} & \sbest{60.2} & \best{84.1}  
                  & \best{14.2} & \best{66.3} & \best{87.1}  
                  & \best{16.0} & \sbest{69.2} & \best{84.1}  
                  & \best{31.3} & \best{22.9} & \best{60.2}  
                  & \sbest{23.1} & \sbest{29.6} & \sbest{74.5} \\
\bottomrule
\end{tabular}
}
\vspace{-4mm}
\end{table*}

We evaluate zero-shot surface normal prediction on standard benchmarks, including NYUv2~\cite{silberman2012indoor}, ScanNet~\cite{dai2017scannet}, iBims-1~\cite{koch2018evaluation}, Sintel~\cite{butler2012naturalistic}, and OASIS~\cite{chen2020oasis}. Normal estimation performance is measured using mean angular error (MAE) and the percentage of pixels with angular errors below 11.25° and 30°. We compare against representative state-of-the-art methods, including Omnidata V2~\cite{kar20223d}, DSINE~\cite{bae2024rethinking}, GeoWizard~\cite{fu2024geowizard}, StableNormal~\cite{ye2024stablenormal}, Diffusion-E2E-FT~\cite{garcia2025fine}, and Lotus-D~\cite{he2024lotus}.

Results in Table~\ref{tab:normal_benchmark} show that our method achieves the best performance on the majority of datasets and evaluation metrics. In particular, our method achieves the lowest mean angular error and the highest accuracy at 30° thresholds on four out of five benchmarks. Even in the few cases where another model slightly surpasses ours, our performance remains among the top three. This demonstrates that our model generalizes robustly across diverse benchmarks.

\subsection{Ablations}

\begin{figure}
\centering
\includegraphics[width=0.95\linewidth]{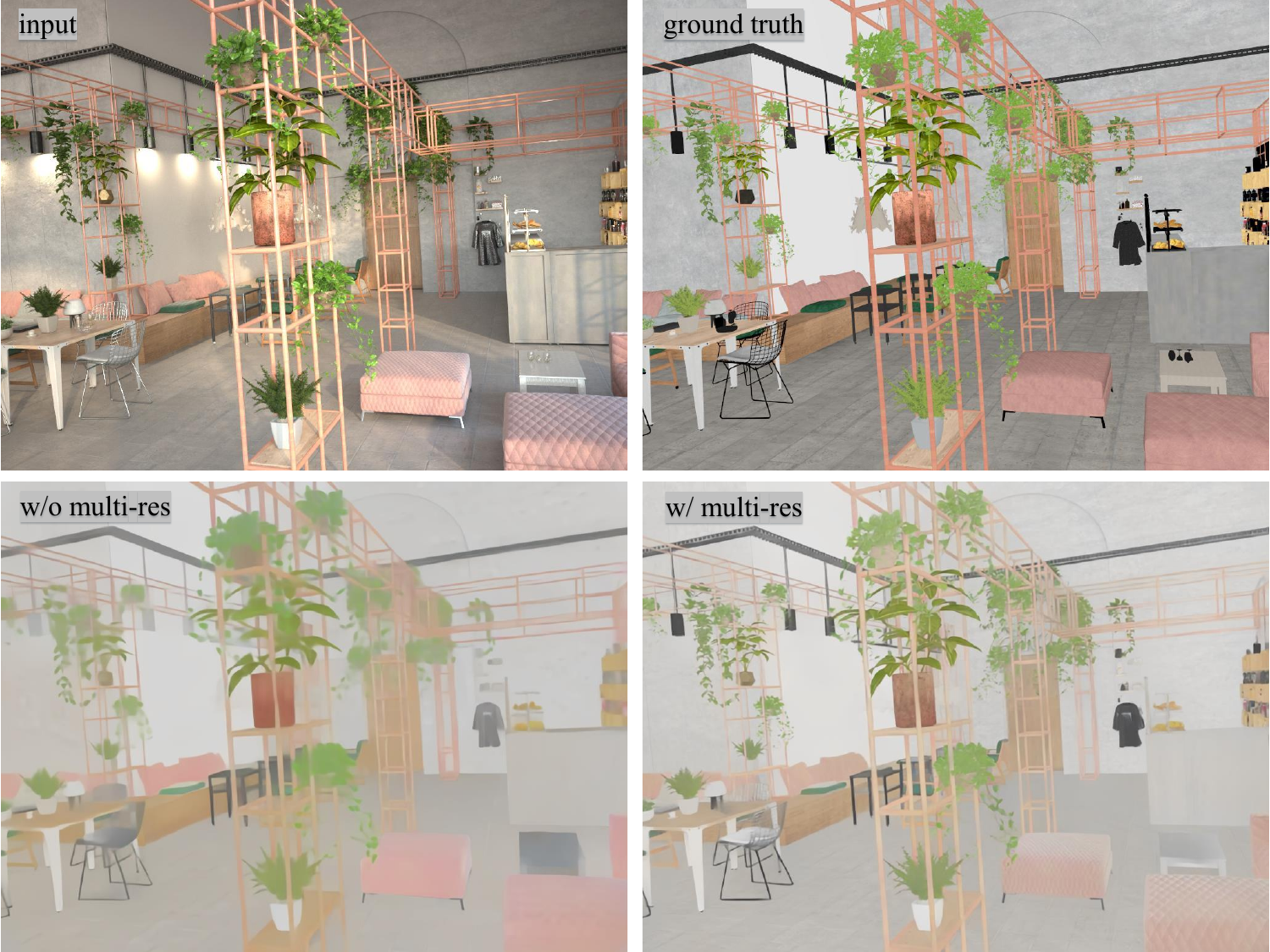}
\caption{\textbf{Effect of Auxiliary Multi-resolution Encoder.} 
Incorporating multi-resolution features from the auxiliary encoder helps recover finer details, yielding more accurate and spatially coherent albedo predictions.}
\label{fig:ablation_res}
\vspace{-2mm}
\end{figure}

\paragraph{Auxiliary multi-resolution encoder preserves fine-grained spatial details.} 
As introduced in Section~\ref{sec:intrinsic_head}, our DPT-style intrinsic heads are augmented with an auxiliary multi-resolution encoder to better preserve high-frequency spatial details. 
This design allows the network to incorporate fine-grained structural cues from the input image without altering the overall architecture. 
As shown in Fig.~\ref{fig:ablation_res}, incorporating the auxiliary encoder leads to more fine-grained and spatially coherent albedo predictions, highlighting the importance of explicitly modeling local details for accurate material recovery.

\paragraph{Consistency finetuning mitigates flickering artifacts.}  
As described in Sec.~\ref{sec:consistency}, we introduce a self-supervised finetuning strategy to enhance temporal stability on real-world videos. 
To quantitatively assess its effectiveness, we evaluate the model before and after finetuning on 100 test videos in the DL3DV~\cite{ling2024dl3dv} dataset by measuring the reprojection error across adjacent frames. As shown in Table~\ref{tab:finetune_ablation}, consistency finetuning notably reduces temporal flickering and yields smoother, more stable material predictions over time.

\begin{table}[t]
\centering
\caption{\textbf{Effect of consistency finetuning.} 
Comparison before and after applying the proposed self-supervised finetuning on real-world videos.}
\resizebox{\columnwidth}{!}{%
\begin{tabular}{lcccc}
\toprule
\multirow{2}{*}{Method} & \multicolumn{4}{c}{RMSE\tdown} \\
\cmidrule(lr){2-5}
 & Albedo & Metallic & Roughness & Shading \\
\midrule
Ours (before finetune) &  0.0341 & 0.0250 & 0.0119 & 0.0150 \\
Ours (after finetune)  & \sbest{0.0161} & \sbest{0.0146} & \sbest{0.0084} & \sbest{0.0140} \\
\bottomrule
\end{tabular}%
}

\label{tab:finetune_ablation}
\vspace{-4mm}
\end{table}

\subsection{Application}

\paragraph{Relighting.}
As shown in the right column of Figure~\ref{fig:teaser}, our approach enables efficient video relighting. Given a sequence of input images, our model predicts view-consistent materials and normals for each frame. With the help of recent feed-forward 3D reconstruction methods~\cite{wang2025pi}, we recover per-frame depth and camera poses to assemble a 3D point cloud of the scene with materials and normals. We then perform physically based rendering (PBR) on this materialized point cloud, allowing novel light sources to be inserted and the entire scene to be relit in real time. The whole pipeline runs for several seconds given the three image inputs, highlighting the efficiency and interactivity of our feed-forward inverse rendering framework. 

\paragraph{View-consistent Image Editing.}
\begin{figure}[t!]
\centering
\includegraphics[width=\linewidth]{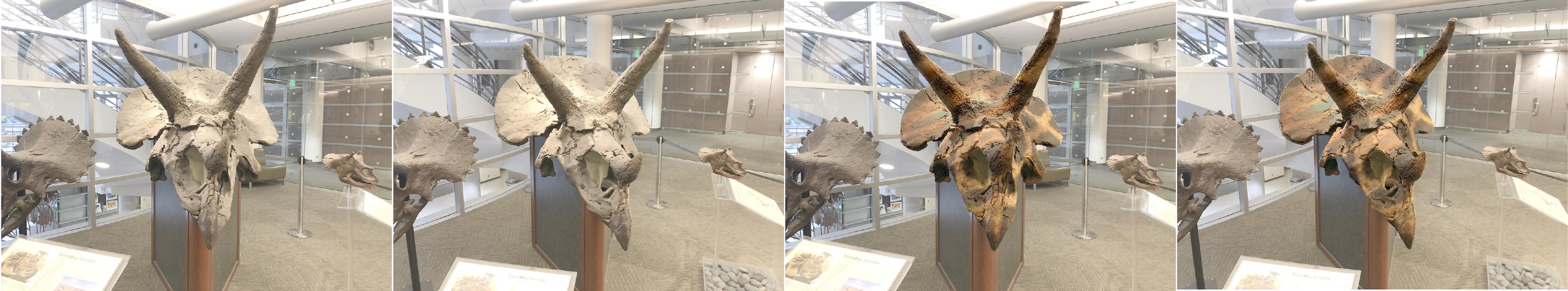}
\caption{\textbf{View-Consistent Material Editing}. 
Our method enables consistent and photorealistic material editing across multiple views. 
Notably, the shadows and shading on the horn are naturally preserved after editing. 
}
\label{fig:material_edit}
\vspace{-4mm}
\end{figure}

With the reconstructed 3D point cloud with PBR materials and normals, we show that we also can perform efficient view-consistent and lighting-aware material editing across multiple input views. Given the region to be modified, we directly replace the albedo of the target points with desired texture and re-render the scene using our predicted diffuse shading. Results can be found in Figure~\ref{fig:material_edit}. 

\section{Conclusion}
In this work, we introduced \textbf{MVInverse}, a feed-forward multi-view inverse rendering framework that achieves fast and consistent intrinsic predictions across input views. By integrating alternating global and frame-wise attention, our method bridges the gap between efficiency and fidelity, enabling efficient material recovery for multi-view captures.
Despite its effectiveness, our approach is still constrained by the limited availability and diversity of material annotations, restricting the model’s ability to generalize across diverse scenes. Future work could focus on collecting more high-quality datasets to validate the architecture’s scalability to further enhance its physical accuracy and generalization to in-the-wild scenes. 
We believe that combining our pipeline with large-scale 3D datasets has the potential to turn feed-forward multi-view inverse rendering into a fundamental component of 3D perception and content creation pipelines.

\clearpage

{
    \small
    \bibliographystyle{ieeenat_fullname}
    \bibliography{main}
}

\clearpage
\setcounter{page}{1}
\maketitlesupplementary

\begin{figure*}[t]
\centering
\includegraphics[width=\linewidth]{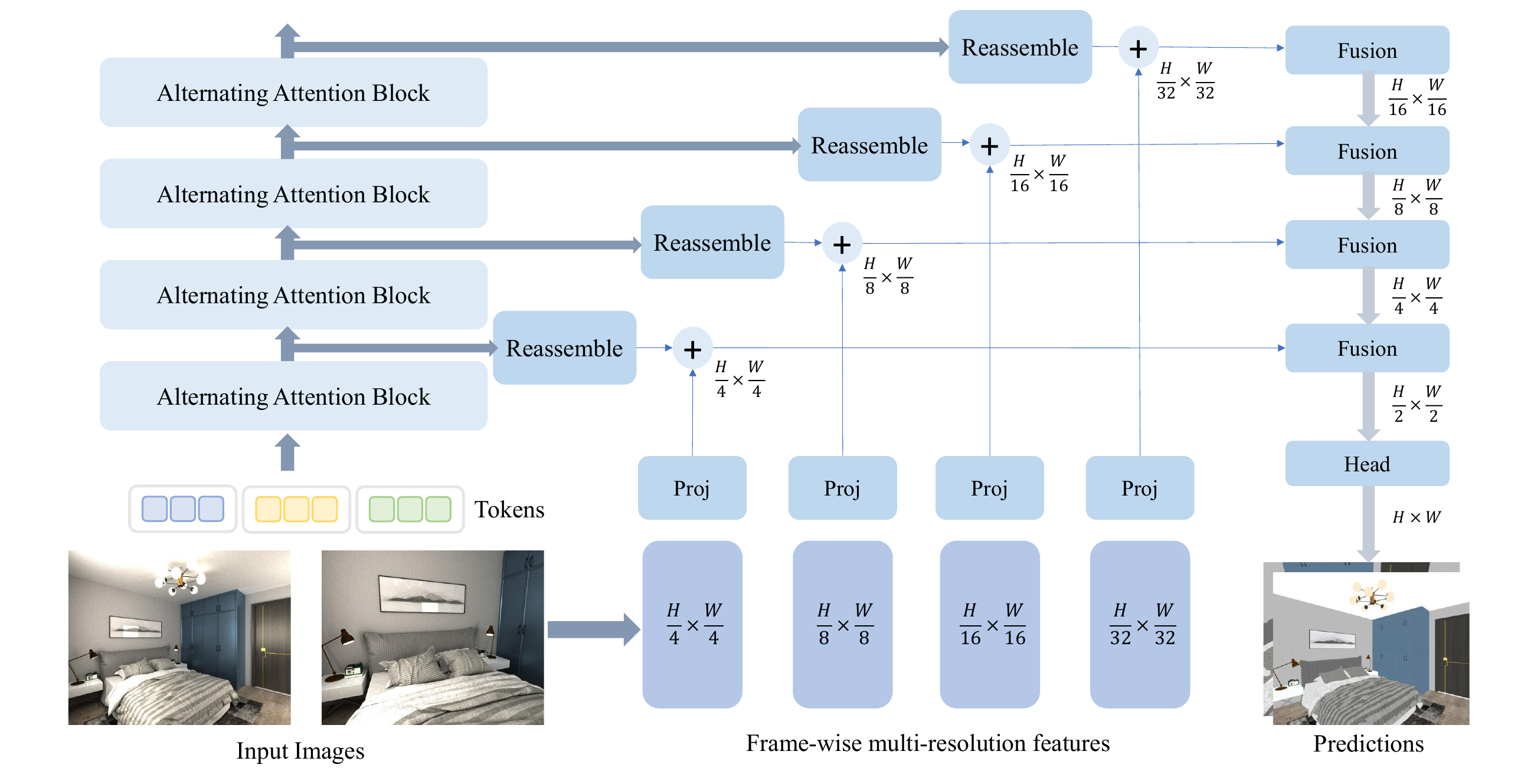}
\vspace{-4mm}
\caption{\textbf{Feature Fusion Detail} 
We illustrate the feature-fusion strategy used to integrate outputs from the encoder stack. Multi-scale features extracted from the DINO encoder (left in the figure) and the ResNeXt backbone (bottom) are projected into a shared embedding space and subsequently fused through a lightweight convolutional refinement block. The \emph{Reassemble} and \emph{Fusion} blocks follow the same design as in~\cite{ranftl2021vision}.
}
\label{fig:supp_feature_fuse}
\end{figure*}
\begin{figure}[h]
\centering
\includegraphics[width=0.95\linewidth]{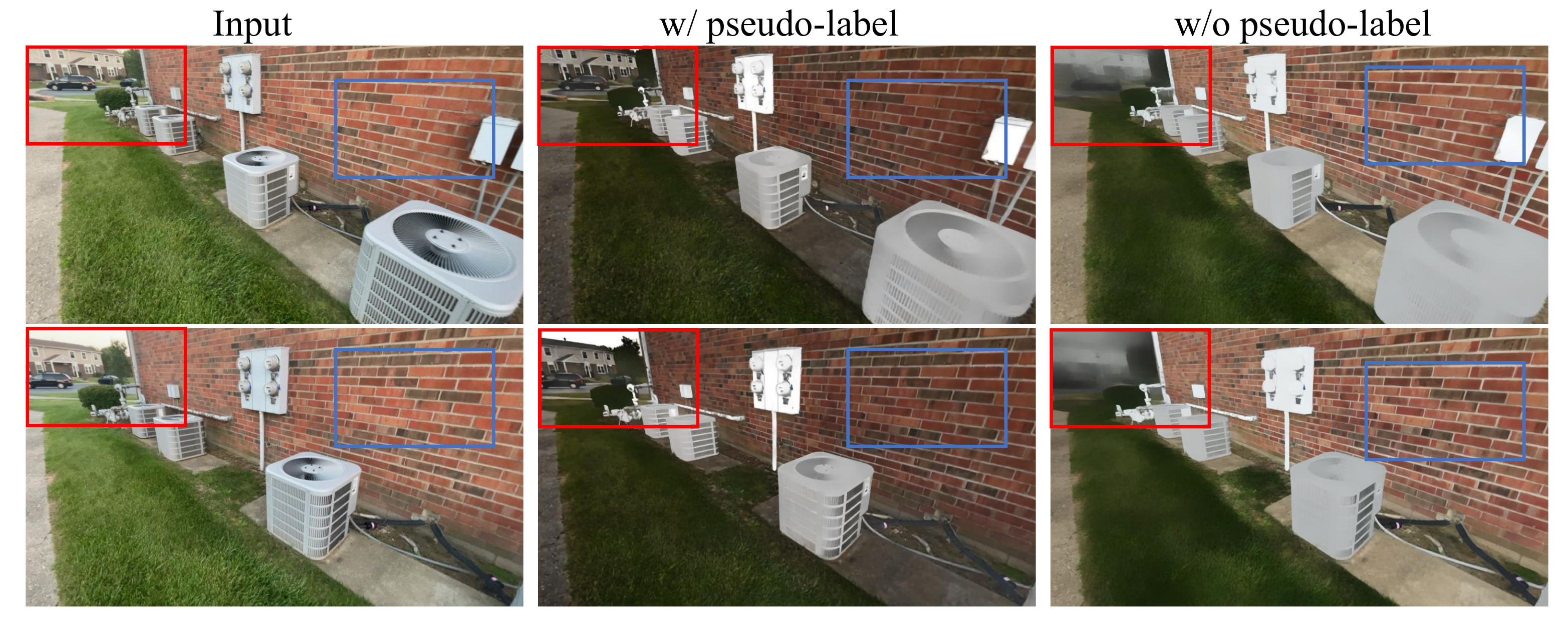}
\caption{\textbf{Impact of pseudo-labels on albedo prediction.}
\textcolor{red}{\textbf{Red}}: Without pseudo-labeled data generated by DiffusionRenderer~\cite{liang2025diffusion}, the model produces blurry patches near sky regions due to the lack of outdoor training samples. Incorporating pseudo-labels significantly improves prediction clarity in these areas.
\textcolor{blue}{\textbf{Blue}}: A side effect of introducing pseudo-labels is a tendency toward darker albedo estimates (e.g., on walls and floors).}
\label{fig:supp_ft_sky}
\vspace{-4mm}
\end{figure}


\begin{figure}[t]
\centering
\includegraphics[width=\linewidth]{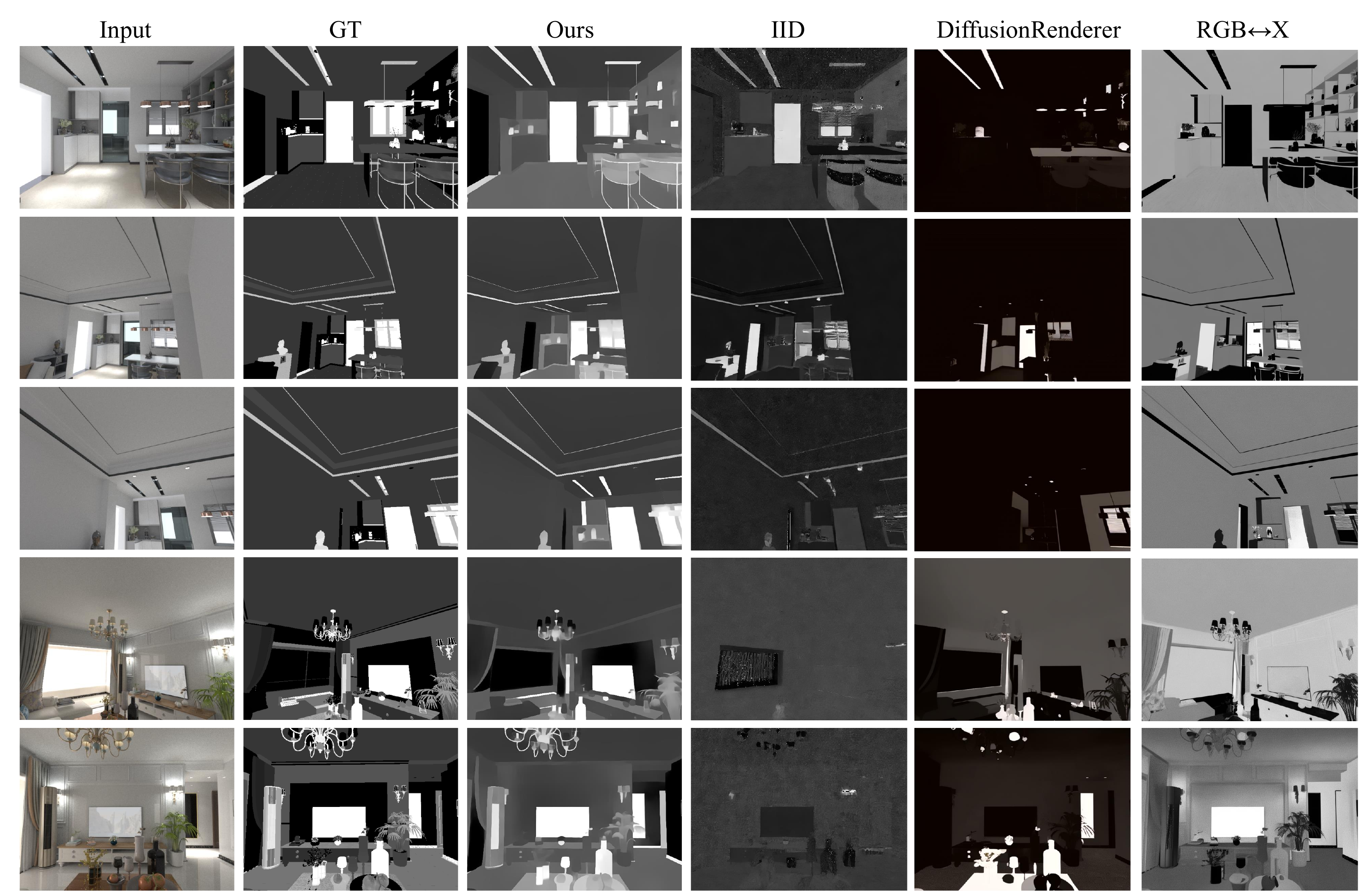}
\caption{\textbf{Qualitative comparison for metallic prediction.} 
}
\label{fig:supp_metallic}
\end{figure}

\begin{figure}[t]
\centering
\includegraphics[width=\linewidth]{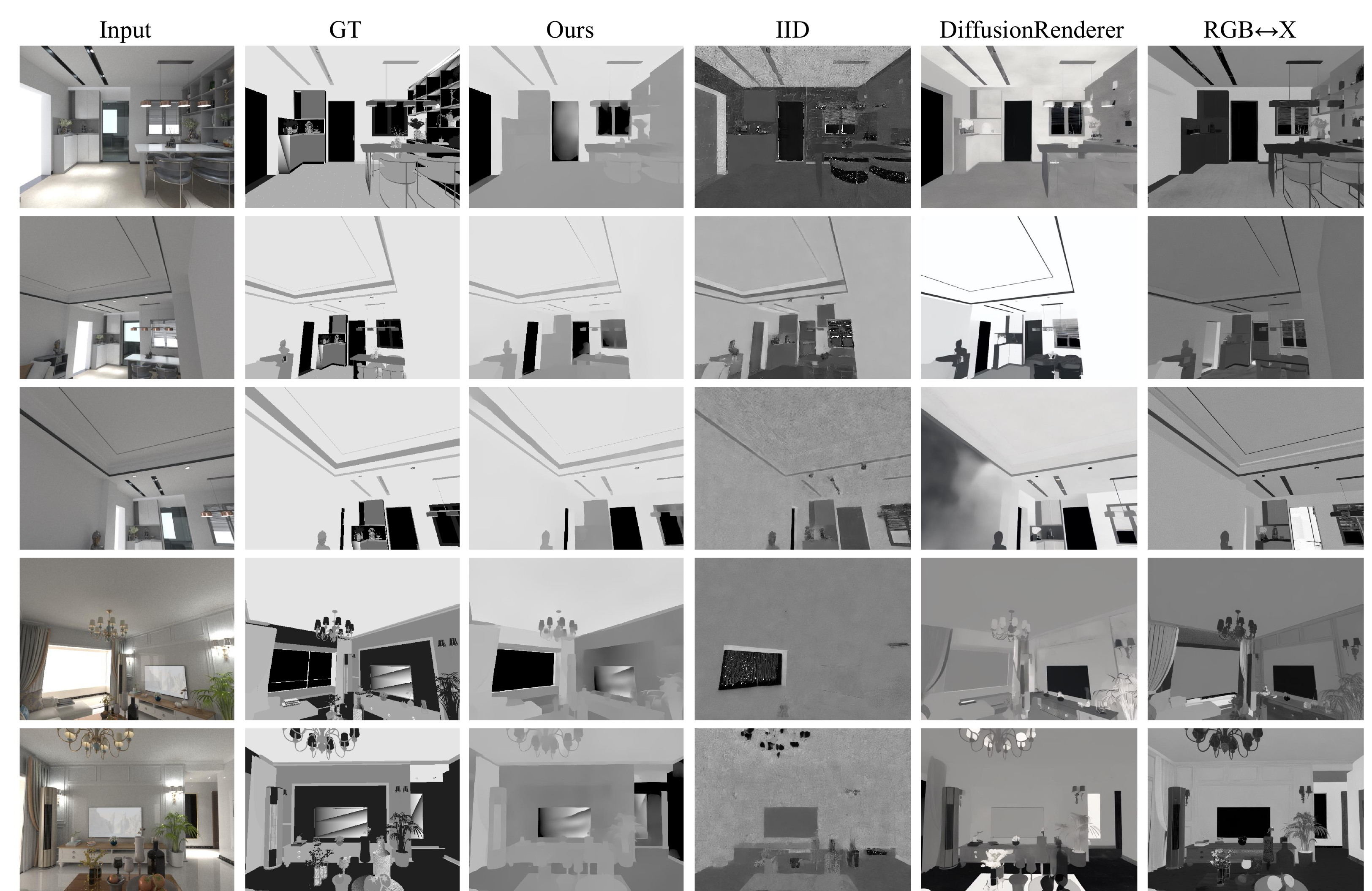}
\caption{\textbf{Qualitative comparison for roughness prediction.} 
}
\label{fig:supp_roughness}
\end{figure}

\section{Architecture Details}

We provide additional details of our network architecture. Our DINO encoder and the alternating-attention transformer are both initialized from the pretrained weights released by Pi3~\cite{wang2025pi}, ensuring stable convergence and strong geometric priors. For the ResNeXt image encoder, we adopt the publicly available weights provided by \cite{wslimageseccv2018}, which offer robust, multi-level representations for detail preservation. Following Pi3~\cite{wang2025pi}, our alternating-attention module contains 18 alternating attention blocks to effectively aggregate multi-view information.

We describe the feature-fusion strategy used to integrate outputs from the encoder stack. As shown in Figure~\ref{fig:supp_feature_fuse}, multi-scale features from the DINO encoder and the ResNeXt backbone are projected to a shared embedding space and fused through a lightweight convolutional refinement block. This design preserves view-dependent details while maintaining geometric consistency across viewpoints.

\section{Training Details}
\begin{table*}[t]
    \centering
    \caption{\textbf{Summary of the dataset.}
    Sekai-Drone~\cite{li2025sekai} uses pseudo-labels generated by DiffusionRenderer~\cite{liang2025diffusion}. 
    For the ABO~\cite{collins2022abo}, we use a randomly selected subset consisting of 100{,}000 images.}
    \label{tab:datasets_summary}
    \begin{tabular*}{\linewidth}{@{\extracolsep{\fill}} l c c l l}
        \toprule
        \textbf{Dataset} & \textbf{\# of Images} & \textbf{View} & \textbf{Scene Type} & \textbf{Intrinsic Types} \\
        \midrule
        Hypersim ~\cite{roberts2021hypersim}       & 70,838  & Multi-view & Indoor & Albedo, Normal, Diffuse Shading \\
        Structured3D ~\cite{zheng2020structured3d} & 78,463  & Multi-view & Indoor & Albedo, Normal \\
        CGIntrinsic ~\cite{li2018cgintrinsics}     & 20,160  & Multi-view & Indoor & Albedo \\
        PRID ~\cite{wang2022high}                  & 21,478  & Single-view & Indoor & Albedo \\
        InteriorVerse ~\cite{zhu2022learning}      & 52,769  & Multi-view & Indoor & Albedo, Metallic, Roughness, Normal \\
        MatrixCity ~\cite{li2023matrixcity}        & 44,804  & Multi-view & Outdoor & Albedo, Metallic, Roughness, Normal \\
        Sekai-Drone$^{*}$ ~\cite{li2025sekai}       & 127,246 & Multi-view & Outdoor & Albedo \\
        ABO$^{*}$ ~\cite{collins2022abo}                  & 100,000 & Multi-view & Object & Albedo, Metallic, Roughness, Normal \\
        \bottomrule
    \end{tabular*}
    \vspace{-4mm}
\end{table*}

In the pretraining stage, we supervise albedo using a scale-invariant reconstruction loss. To account for the scale ambiguity in albedo prediction, we first estimate a per-channel scale factor that aligns the prediction to the pseudo-ground-truth using least-squares error minimization. Specifically, for a predicted albedo map $A \in \mathbb{R}^{H\times W \times 3}$ and 
ground-truth $A^{*}$, we compute the 3-channel scale factor as
\begin{equation}
    s^\ast = \arg\min_{s \in \mathbb{R}^3} \|A\odot s - A^{*}\|_2^2 ,
\end{equation}
where $s$ is a 3D scale vector and $\odot$ denotes channel-wise multiplication. 
The scale-invariant albedo loss is then defined as
\begin{equation}
    \mathcal{L}_{\text{mse}} = \frac{1}{N}\|A\odot s^\ast - A\|_2^2 .
\end{equation}
Since this loss is unstable at early iterations, we first warm up the network using a vanilla MSE loss for several epochs before switching to the scale-invariant formulation, following ~\cite{careagaColorful, li2018cgintrinsics}. 

Throughout the pretraining stage, both the DINO encoder and ResNeXt encoder remain frozen to preserve their pretrained representations. To accommodate varying input frame counts, we adopt a dynamic batch sizing strategy similar to VGGT. Each GPU processes up to 12 images, and each batch is formed by randomly sampling 2 to 12 images of the same scene. For datasets that provide only single-view images (e.g., PRID~\cite{wang2022high}), we replicate each monocular image along the batch dimension to match the required number of input frames, effectively treating them as multiple captures from a static camera.  We train for 80 epochs on two A100 GPUs at a fixed long-side resolution of 518 (with the short side randomly sampled), with each epoch consisting of 1000 iterations.After this initial training, we freeze all attention blocks and continue training only the prediction heads at a long-side resolution of 770 for an additional 50 epochs. This fine-tuning stage focuses on improving high-frequency details while keeping the reasoning in backbones fixed. The full training pipeline requires approximately 3–4 days. For all stages, we use the Adam optimizer with an initial learning rate of 
$5 \times 10^{-5}$. 

Table~\ref{tab:datasets_summary} summarizes all datasets and their corresponding statistics. While most of our benchmarks consist of indoor environments, we additionally include two outdoor datasets—MatrixCity~\cite{li2023matrixcity} and Sekai (for which we use pseudo-labels generated by DiffusionRenderer)—as well as the object-centric ABO dataset~\cite{collins2022abo}. Following the masking strategy used in~\cite{careagaColorful}, we apply masks to exclude unreliable regions during supervision. Specifically, if a dataset provides masks for specular or mirror surfaces, we use these masks and compute losses only within the masked regions. In the absence of such annotations, we instead mask out pixels whose albedo values fall below 0.01 or above 0.99 to prevent these non-informative values from affecting optimization.


\section{More Qualitative Results}
\begin{figure}[t]
\centering
\includegraphics[width=\linewidth]{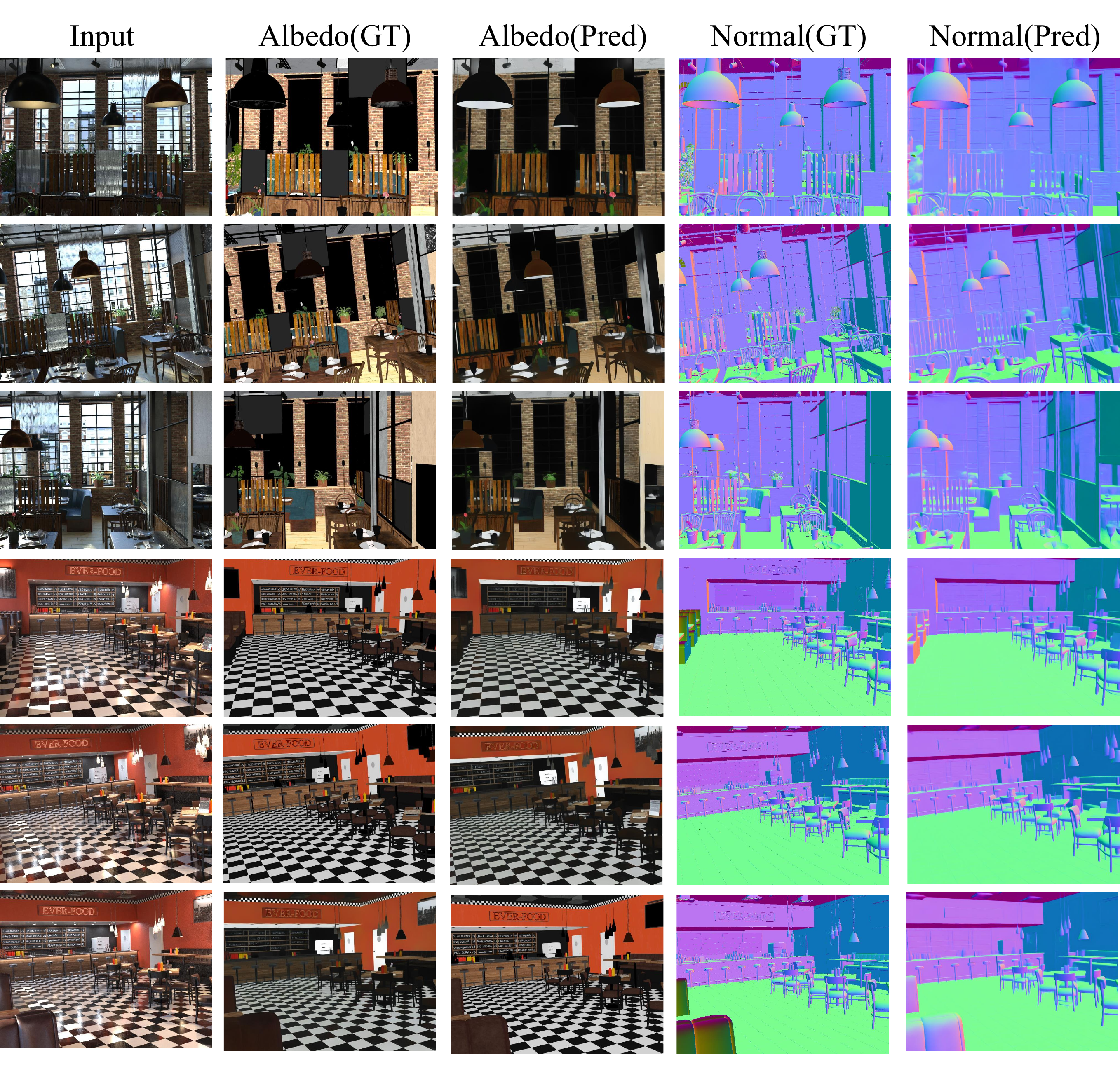}
\caption{\textbf{Qualitative comparison on in-distribution Hypersim~\cite{roberts2021hypersim} dataset.} 
}
\label{fig:supp_hypersim_1}
\vspace{-4mm}
\end{figure}

\begin{figure}[t]
\centering
\includegraphics[width=\linewidth]{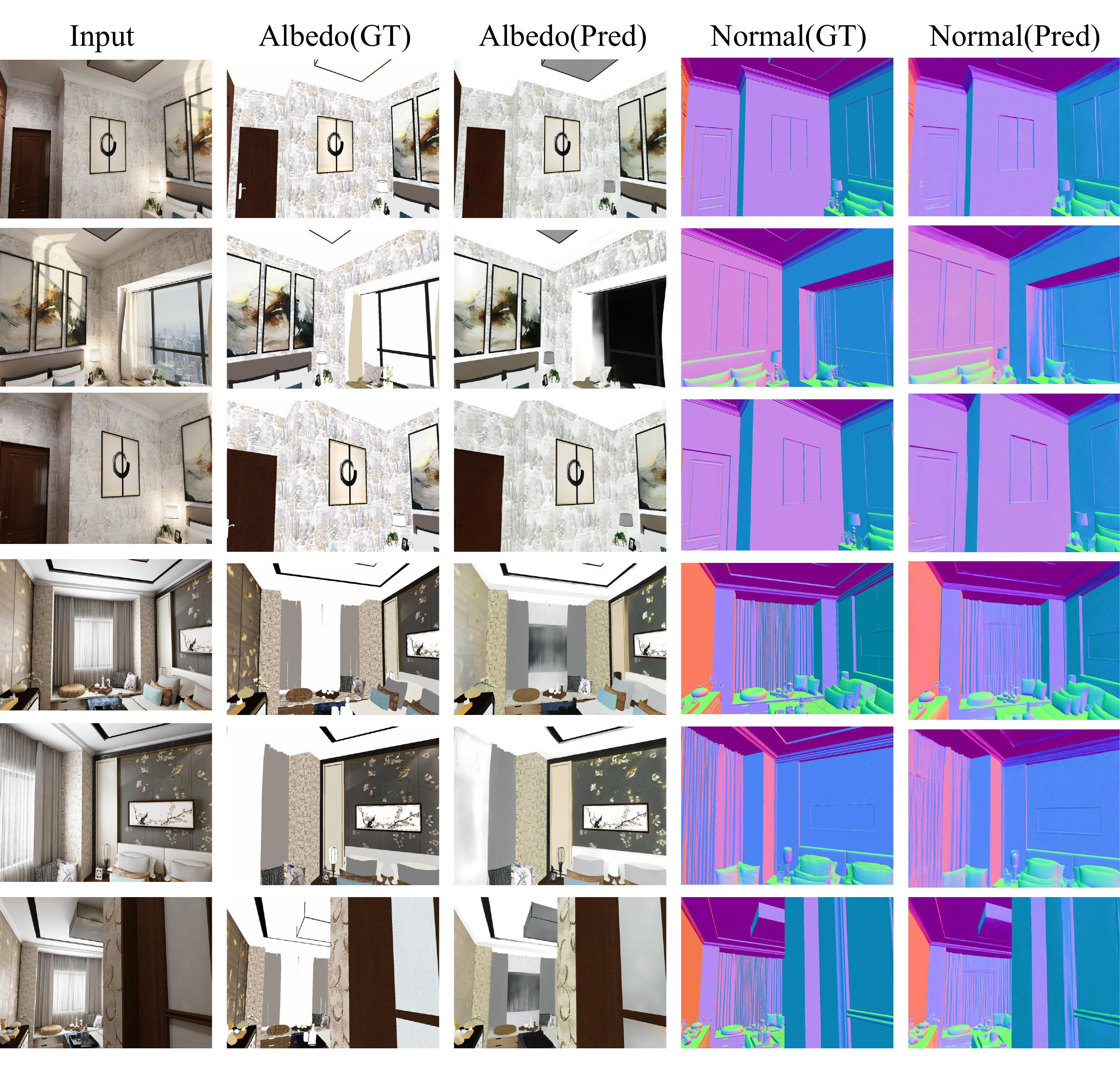}
\caption{\textbf{Qualitative comparison on in-distribution Structured3D~\cite{zheng2020structured3d} dataset.} 
}
\label{fig:supp_structured3d}
\vspace{-8mm}
\end{figure}

\begin{figure*}[t!]
\centering
\includegraphics[width=0.95\linewidth]{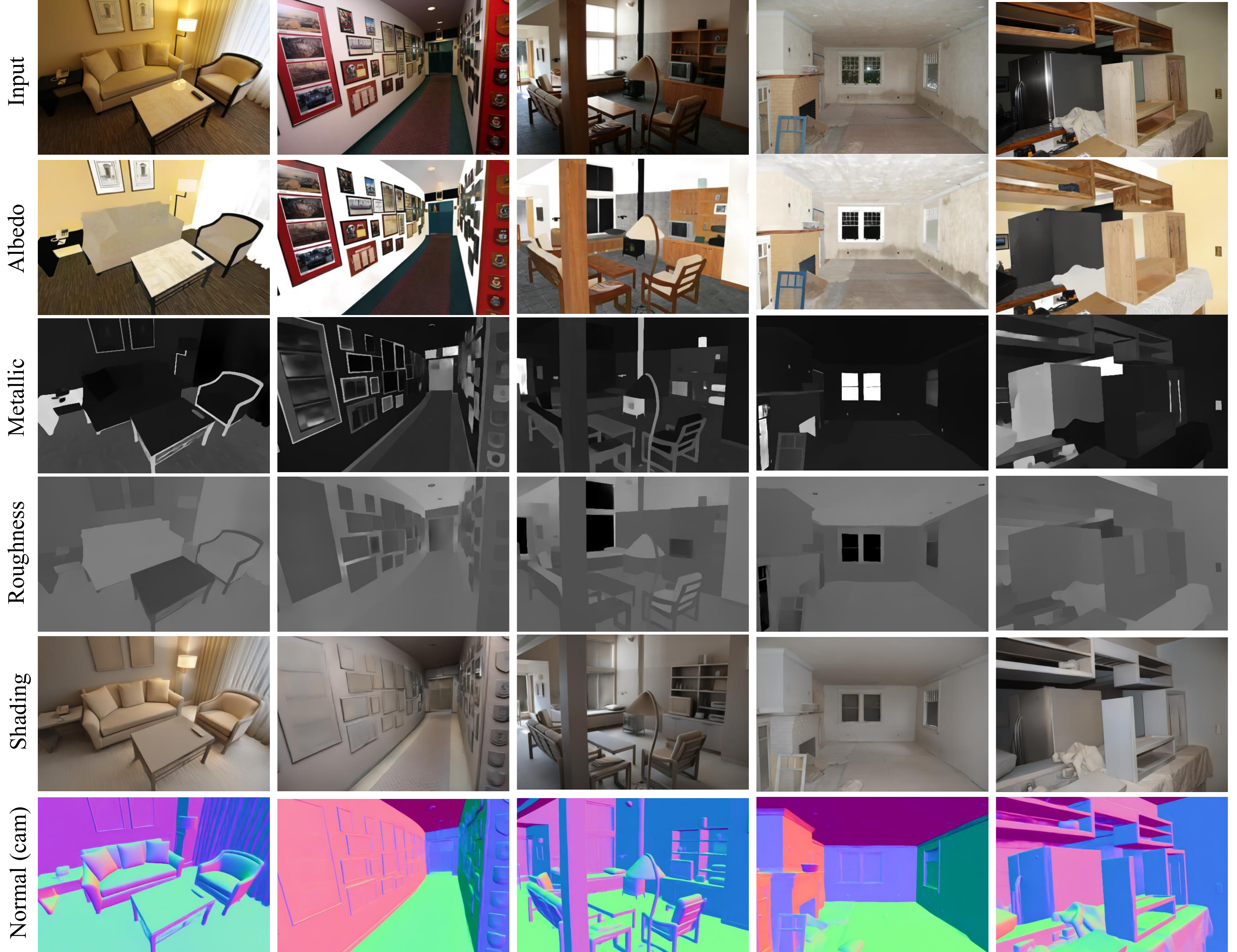}
\caption{\textbf{More single-view results on IIW} 
We show additional albedo predictions on the IIW dataset to highlight the ability of our model to predict accurate and visually pleasing intrinsic images, including albedo, metallic, roughness, diffuse shading and camera-space normals. 
}
\label{fig:supp_iiw}
\end{figure*}
\begin{figure*}[t]
\vspace{-9mm}
\centering
\includegraphics[width=0.95\linewidth]{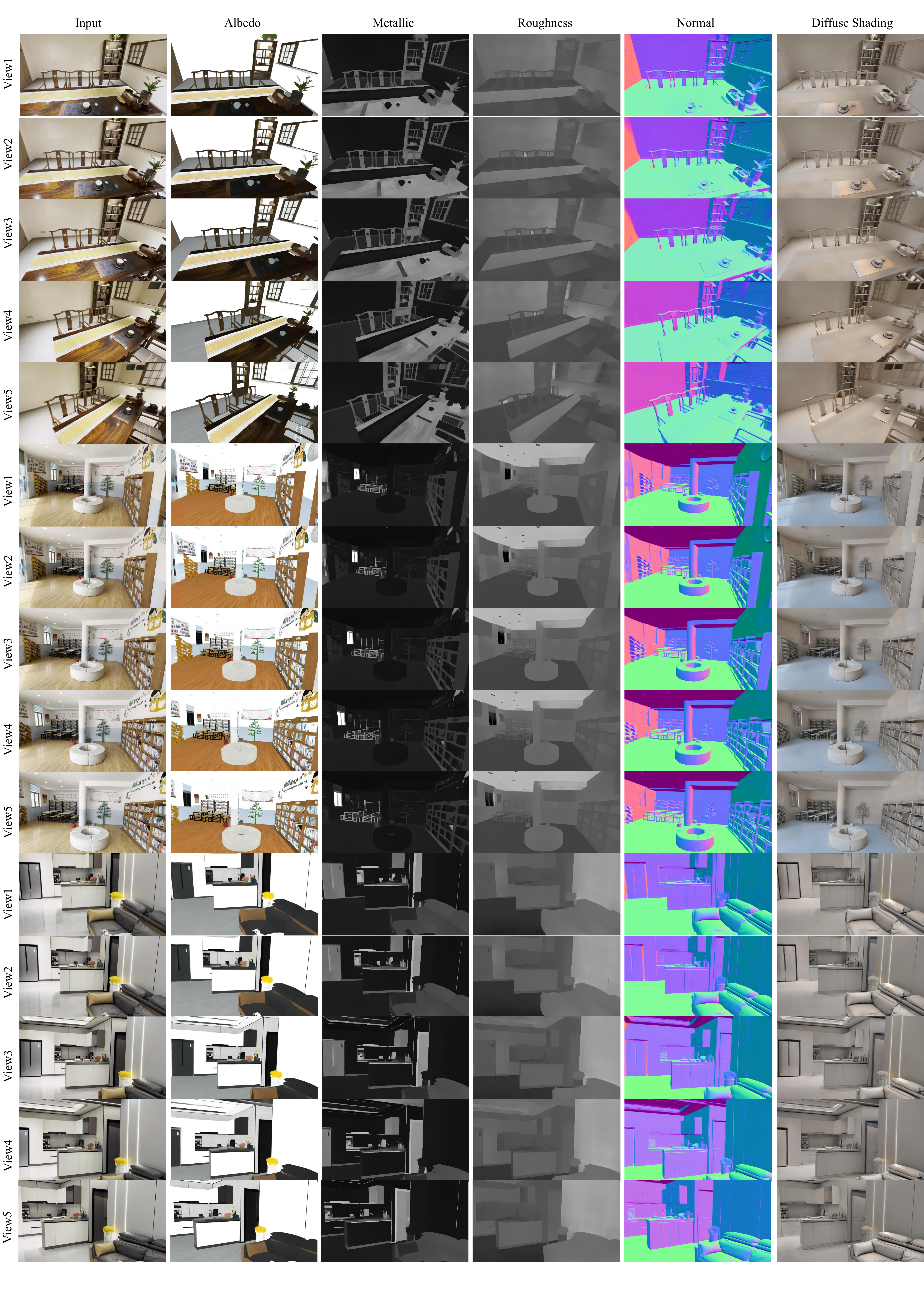}
\vspace{-5mm}
\caption{\textbf{Multi-view results on the DL3DV~\cite{ling2024dl3dv} dataset} demonstrating the cross-view consistency and generalization of our method.}
\label{fig:supp_mv}
\end{figure*}

In this section, we provide additional qualitative results to further demonstrate the effectiveness, robustness, and generalization capability of our feed-forward inverse rendering model.

\paragraph{Single-view Prediction.}
Figure~\ref{fig:supp_metallic},~\ref{fig:supp_roughness} presents supplementary single-view predictions of metallic and roughness maps on the InteriorVerse~\cite{zhu2022learning} dataset. 
Since these material channels are not qualitatively visualized in the main paper, we include extended comparisons here against three recent state-of-the-art baselines---RGB$\leftrightarrow$X~\cite{zeng2024rgb}, IntrinsicImageDiffusion~\cite{kocsis2024intrinsic}, and DiffusionRenderer~\cite{liang2025diffusion}. 
As shown in the figures, our predictions are closer to ground truth compared to other methods.

We additionally provide more single-view prediction results on the test set of Hypersim~\cite{roberts2021hypersim} and Structured3D~\cite{zheng2020structured3d} in Figure~\ref{fig:supp_hypersim_1}, ~\ref{fig:supp_structured3d}. We also provide in the wild results in IIW~\cite{bell2014intrinsic} dataset in Figure~\ref{fig:supp_iiw}. 
These examples highlight our model's ability to generalize to diverse synthetic or real-world scenes. 


\paragraph{Multi-view Prediction.}
Figure~\ref{fig:supp_mv} show multi-view prediction results on the DL3DV~\cite{ling2024dl3dv} dataset, which serves as an out-of-distribution (OOD) benchmark due to its diverse large-scale scenes and complex geometry. 
For each multi-view input sequence, we visualize the predicted albedo, metallic, roughness, camera-space normals, and diffuse shading. 
Across all examples, our method demonstrates strong cross-view consistency and stable material predictions. 

\section{Discussions and Limitations}


\paragraph{Impact of pseudo-labeled data.} We leveraged a subset of real-world pseudo-labels generated by DiffusionRenderer to train our albedo model, aiming to improve generalization in real-world outdoor scenes. Here we discuss the impact of pseudo-labels. As illustrated in Figure~\ref{fig:supp_ft_sky}, incorporating pseudo-labels significantly enhances prediction quality in challenging regions, such as near-sky areas, which are otherwise underrepresented in our indoor-dominated training data. Specifically, without pseudo-labels, the model tends to produce blurry or inconsistent albedo estimates near sky regions due to the lack of outdoor supervision. The introduction of pseudo-labels mitigates this issue, resulting in sharper and more physically plausible predictions. However, this approach also has some drawbacks: the model exhibits a tendency toward darker albedo estimates in certain regions, such as walls and floors, which results from the sub-optimal quality of the pseudo-labels. While the use of pseudo-labels is a pragmatic solution given the limited availability of annotated outdoor data, it remains a compromise. To further improve generalization to outdoor environments, a more comprehensive dataset covering diverse outdoor distributions would be required.


\paragraph{Performance Limitations.} 
\begin{figure*}[t]
\centering
\includegraphics[width=0.95\linewidth]{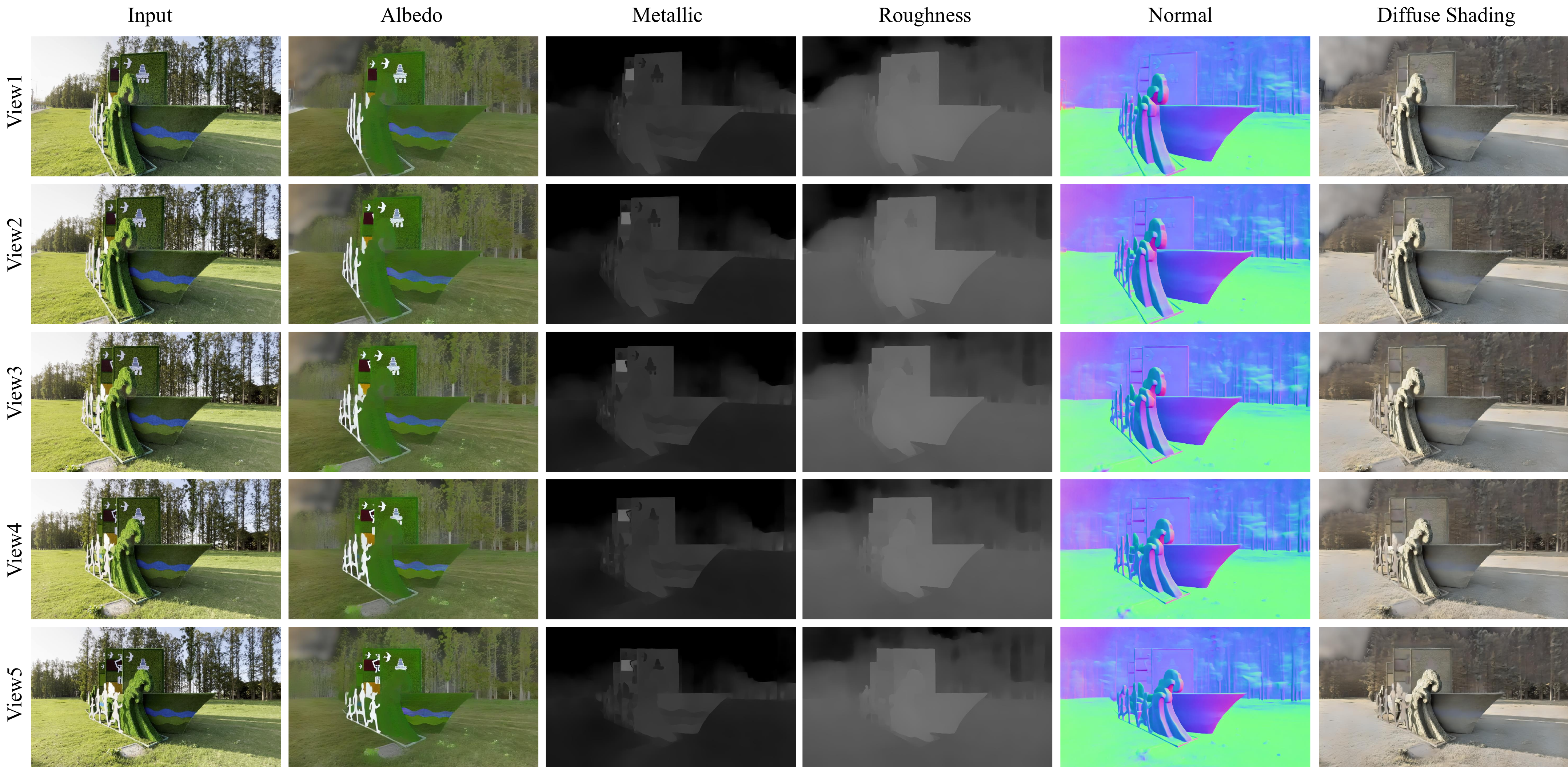}
\caption{\textbf{Another example on the DL3DV~\cite{ling2024dl3dv} dataset} demonstrating its limitation when applied to outdoor scenes.}
\label{fig:supp_mv_5}
\end{figure*}
As our method is fundamentally data-driven, its performance can be constrained by the distribution of the training data when applied to out-of-distribution (OOD) inputs. This occasionally leads to predictions with inaccurate chromaticity. For example, in the top-most scene in Figure~\ref{fig:supp_mv}, the albedo of the wooden table is predicted darker than expected, while in the middle scene, the predicted albedo of the wooden floor exhibits noticeable tonal variations. However, it is important to note that such behavior is a common limitation of data-driven approaches; for instance, Figure.5 in the main paper demonstrates a similar drawback of DiffusionRenderer (the picture with a white cat on the floor). Beyond expanding the diversity of training data, this issue might also be mitigated by explicitly incorporating chromaticity information into the albedo prediction process, following the approach of~\cite{careagaColorful}.

Moreover, labeled data for metallic and roughness predictions are much sparser compared to albedo, with annotations available only in a few datasets such as MatrixCity, InteriorVerse, and ABO. This scarcity significantly limits the model’s generalization capability, leading to inconsistent or unreliable predictions for these material properties. For example, the model rarely encounters trees with ground-truth metallic or roughness labels, resulting in blurry and questionable predictions in these areas, as illustrated in Figure~\ref{fig:supp_mv_5}.

In addition, some albedo datasets exhibit inherently unreliable labels. For instance, in Hypersim~\cite{roberts2021hypersim}, specular and mirror surfaces are assigned near-zero albedo values, whereas in Structured3D~\cite{zheng2020structured3d}, such surfaces instead receive unrealistically bright albedo. This ambiguity can lead to ambiguous, grayish outputs (see the window area in bottom row of Figure~\ref{fig:supp_structured3d}). This arises because our model is discriminative: when faced with inconsistent supervision, it tends to regress toward median values within these regions, thereby producing blurred results. Although we mask out pixels with albedo values below 0.01 or above 0.99, the model may still learn to produce overly dark or gray predictions around specular or mirror surfaces. A potential way to mitigate this issue is to incorporate datasets with more reliable albedo annotations or more consistent treatment of specular/glass surfaces.

\end{document}